\documentclass[11pt]{article}
\usepackage[T1]{fontenc}
\usepackage{amsmath,amssymb,graphicx,url}

\usepackage[draft]{changes} 

\usepackage{algorithm}

\usepackage{svg}
\usepackage{booktabs}
\usepackage{array}
\usepackage{makecell}
\usepackage{multirow}
\usepackage{caption}
\usepackage[margin=1in]{geometry}
\usepackage{lmodern}

\usepackage{booktabs,tabularx,array}

\usepackage{algpseudocode}

\usepackage[colorlinks=true,citecolor=blue,linkcolor=black,urlcolor=black]{hyperref}
\usepackage{bookmark}

\newcommand{\ImgRefPrompt}{%
\textit{Classify the wound shown in the query image.\newline
Choose exactly one label from: \{class labels\}.\newline
Use the exact label spelling.\newline
Reply with exactly one label name and nothing else.}}

\title{In-Context Learning for Wound Classification with Small Multimodal Language Models}

\author{
George Martvel$^{1@}$, Oskar Gustafsson$^{2,3}$, John Pavia$^{3}$, Ernst Ahlberg$^{3,4}$\\[0.5em]%, Lars Carlsson$^{1,4}$\\[0.5em]
\small $^{1}$Jönköping University, Jönköping, Sweden\\
\small $^{2}$Halmstad University, Halmstad, Sweden\\
\small $^{3}$Mölnlycke Health Care\\
\small $^{4}$Centre for Reliable Machine Learning, Royal Holloway, University of London, Egham, Surrey, UK\\[0.55em]
\small $^{@}$Corresponding author: georgy.martvel@ju.se}

\date{}

\begin{document}

\maketitle

\begin{abstract}

\textbf{Background and Objective:} Wound image classification is often treated as a task-specific supervised learning problem, requiring substantial amounts of manually labelled data and retraining when the label space or deployment setting changes. This study evaluated whether small multimodal language models (SMLMs) can provide a training-free alternative for wound classification through retrieval-based in-context learning (ICL).

\textbf{Methods:} Experiments used two public wound-image datasets: the Kaggle wound dataset (1469 images, 10 classes) and the Medetec dataset (560 images, 9 classes). Eleven SMLMs from the Qwen 3.5, Ministral 3, and Gemma 4 families were evaluated under zero-shot prompting and few-shot prompting with random support examples, embedding-based $k$-nearest-neighbour ($k$NN) retrieval, and $k$NN retrieval followed by maximal marginal relevance reranking (MMR). Retrieval-only weighted-$k$NN controls, support-set reduction experiments, and support-context size sweeps were used to assess the effects of retrieval, model scale, and prompt length.

\textbf{Results:} Query-conditioned ICL consistently outperformed zero-shot and random few-shot prompting. On the Kaggle dataset, the best result was achieved by Qwen 3.5 27B with $k$NN+MMR, reaching 0.872 accuracy and 0.871 F1 score. On Medetec, Qwen 3.5 27B with $k$NN+MMR reached 0.678 accuracy and 0.670 F1. Larger models exceeded matched weighted-$k$NN controls, indicating use of retrieved examples beyond nearest-neighbour voting. Retrieval-based ICL degraded modestly under support-set reduction, and most gains saturated with 8–10 support images.

\textbf{Conclusions:} Retrieval-based ICL allows SMLMs to perform adaptable wound image classification without task-specific retraining. Compact retrieved contexts may support practical and privacy-conscious deployment, although performance remains dependent on model scale, retrieval strategy, and dataset difficulty.

\end{abstract}

\section{Introduction}

Computer vision is playing an increasingly important role in medical imaging~\cite{aggarwal2021diagnostic}, yet many high-performing systems still follow a task-specific paradigm: a dataset is collected and annotated for a single objective, a dedicated model is trained or fine-tuned, and the process is repeated whenever the task definition, label set, or deployment context changes~\cite{bian2025artificial, guan2021domain}. Although this approach can yield strong performance, it also requires substantial annotation effort and computational resources, particularly in settings where expert labels are scarce or practical requirements evolve rapidly~\cite{wang2024comprehensive}. Recent advances in large multimodal language models (LMLMs) suggest an alternative to conventional supervised learning~\cite{bradshaw2025large, nam2025multimodal, sun2025visual, tsimpoukelli2021multimodal}. Modern LMLMs can be applied in a zero-shot setting, in which the model receives only task instructions at inference time and requires neither annotated training data nor task-specific fine-tuning~\cite{li2023blip,liu2023visual}. In practice, however, this approach often yields moderate to good performance, depending on the complexity and specificity of the task~\cite{jiang2024many,nam2025multimodal, sun2025medical}. 

The performance of LMLMs on specific tasks can be improved through adaptation strategies such as fine-tuning and in-context learning (ICL)~\cite{alayrac2022flamingo, brown2020language, liu2023visual}. In fine-tuning, model parameters are updated using task-specific training data, either across the full model or through lighter-weight adaptation schemes, which can yield strong performance when sufficient annotations and computational resources are available~\cite{dettmers2023qlora, hu2022lora}. In contrast, in ICL, a small number of labelled examples are provided directly in the prompt at inference time rather than being learned through training~\cite{alayrac2022flamingo, dong2024survey}. Such prompting-based approaches are especially appealing in medicine because they allow for rapid adaptation with minimal task-specific data and without requiring a separate training pipeline for each application~\cite{moor2023med, nori2023can}. %Early evidence shows that zero‑shot and few‑shot prompting can be effective in medical image analysis, with ICL improving GPT-4V performance on cancer histopathology tasks and, in some settings, matching or exceeding specialised classifiers~\cite{ferber2024context}. Similarly, in diabetic retinopathy detection, Gemini 1.5 Pro with ICL achieved accuracy and F1 scores comparable to a fine-tuned domain-specific model, although it showed worse performance on calibration metrics~\cite{ayhan2025context}.

%At the same time, the benefit of image-based ICL depends strongly on the examples included in the prompt. Prior studies show that retrieval-based selection (such as semantically close images) can outperform random selection, and that ICL design choices such as retrieval strategy, example order, and prompt format can influence LMLM performance~\cite{chen2025can,li2024configure,luo2024does, zhang2023makes}. Nevertheless, similar image (nearest-neighbour) selection remains only a partial solution: the selected examples may be redundant, or similar to the query for reasons that are not relevant to the target distinction~\cite{lee2026learning, suo2024visual}. Counterfactual and diversity-aware retrieval approaches address this by balancing query relevance with diversity, attribute coverage, and contrastive or counterfactual information, rather than relying solely on nearest-neighbour similarity~\cite{kapuriya2025exploring, xiong2026retrieving}.

A further practical issue is the choice of model scale. In deployment settings, one often distinguishes between smaller models that can run locally on consumer or edge hardware and larger models that typically require server‑side or cloud inference. This distinction is only approximate, since the boundary is continuously shifting with developments in hardware, model architecture, and optimisation methods such as quantisation, pruning, and distillation~\cite{lu2024small,van2025survey,xu2024device}. In this work, we use the term small multimodal language models (SMLMs) in a deployment‑oriented sense, referring to models suitable for local or edge‑based inference rather than defining the term by a fixed parameter threshold. This deployment distinction is closely linked to an expected performance tradeoff: scaling-law studies show that, within comparable model families, performance improves predictably with model size, data, and training compute. However, size alone does not determine performance, and smaller models can be competitive when they are better trained or optimised~\cite{hoffmann2022training,kaplan2020scaling}. Thus, even though larger models are attractive when maximum task performance is the main objective, the dependence on server-side inference can be difficult to justify in privacy-sensitive settings, such as internal documentation, governmental workflows, or medical records, where data cannot always be sent to external model providers because of confidentiality, regulatory, or data-governance constraints~\cite{agarwal2024prompt, barbera2025ai, zhong2025considerations}. Locally deployable SMLMs are therefore a realistic alternative~\cite{garg2025rise}, but their usefulness depends on how much performance is lost relative to larger remote models, and whether that loss is acceptable for the intended application.

These concerns are particularly relevant for wound analysis and classification. Wound images are clinically sensitive, visually heterogeneous, and often collected under variable lighting, distance, body location, and image-quality conditions. Additionally, labels usually require wound-care expertise and may differ across classification tasks such as wound type, tissue composition, infection, or referral urgency~\cite{anisuzzaman2022image,tabjabortesi2024machine, zhang2022survey}. This makes the combination of SMLMs and ICL especially attractive: local inference can reduce the need to transmit patient images to external servers, while in-context examples can specify the target label set and decision criteria at inference time without retraining a separate model for every wound-related task~\cite{ayhan2025context,ferber2024context,garg2025rise, moor2023med}. In such a setting, the goal is not necessarily to show that SMLMs outperform or match the performance of larger remote models, but to test whether example-guided local models can provide a useful privacy-preserving compromise between task adaptability, computational cost, and classification performance.

In this work, we examine whether small multimodal language models can serve as a practical solution for wound classification when task-specific training data is limited. Using open wound-image datasets, we evaluate zero-shot and few-shot prompting across eleven models from the Qwen~3.5~\cite{qwen}, Ministral~3~\cite{mistral2026ministral3}, and Gemma~4~\cite{gemma_2026_g4} families, covering a range of SMLM scales under a common prompting protocol. We compare three strategies for constructing the few-shot context: random support sampling, embedding-based $k$-nearest-neighbour ($k$NN) retrieval, and $k$NN retrieval followed by maximal marginal relevance reranking (MMR). To separate the contribution of in-context learning from that of retrieval itself, we also compare each retrieval-augmented prompting condition against a matched retrieval-only weighted-$k$NN baseline. Finally, we examine how progressively reducing the support/training set degrades the accuracy of retrieval-augmented prompting methods and conventional supervised classifiers, and evaluate how the size of the support context (number of support images) affects the performance of SMLMs.

\section{Related Work}

Prior research on wound image analysis has predominantly framed the task as a supervised computer vision problem, focusing on wound classification, detection, segmentation, tissue analysis, and measurement, usually with models trained for a specific wound-related endpoint and evaluated on a fixed label space~\cite{anisuzzaman2022image,wang2015unified,yadav2019feature,zhang2022survey}. Earlier studies typically employed handcrafted colour, texture, or deep-feature representations with conventional classifiers~\cite{abubakar2019can, yadav2019feature}, while later work has moved toward CNN-based and deep ensemble methods for tasks such as infection, ischaemia, or wound-type classification~\cite{aldoulah2023novel, goyal2020recognition,rostami2021multiclass}. A parallel line of work focuses on wound localisation, segmentation, and measurement, including tissue segmentation ~\cite{ramachandram2022fully, scebba2022detect,wang2015unified,wang2020fully}. Multimodal wound classification models have also been proposed, for instance by combining wound images with anatomical location information; however, these approaches still operate within a supervised learning framework and require training for predefined tasks~\cite{anisuzzaman2022multimodal,patel2024integrated}. Similarly, work on surgical wound infection assessment and postoperative monitoring has mostly focused on models designed for a specific clinical endpoint, such as identifying infection risk from wound images, patient-reported symptoms, or other structured inputs~\cite{tabjabortesi2024machine, mclean2025multimodal,wu2020unified}. Overall, the existing literature demonstrates that image-based wound assessment is an active research area, but it remains dominated by task-specific supervised systems.

Large language models and foundation models provide a possible way to move beyond fully task-specific wound-analysis pipelines. In medical imaging, recent surveys discuss pretrained models as reusable backbones for transfer learning, domain adaptation, and annotation-efficient learning~\cite{azad2023foundational,zhang2024challenges}. Some studies have tested whether general-purpose visual representations transfer to medical images: for example, DINOv2~\cite{oquab2023dinov2} has been evaluated on radiology benchmarks using settings such as $k$NN classification, few-shot learning, linear probing, and fine-tuning, showing that natural-image pretraining can provide useful features for some medical tasks~\cite{baharoon2023evaluating}. Vision–language models represent a significant development in this area, as exemplified by CLIP, which introduced shared image–text embeddings that support zero-shot classification and image retrieval~\cite{radford2021clip}. Biomedical variants of CLIP, such as PubMedCLIP, MedCLIP, and BiomedCLIP adapt this idea to medical image-text data~\cite{eslami2023pubmedclip,wang2022medclip,zhang2023biomedclip}. For wound classification, this suggests that both general and medical-pretrained image or image-text representations may be useful when wound-specific annotations are limited.

A related line of work investigates whether pretrained multimodal models can be adapted through examples at inference time (in-context learning). Med‑Flamingo was one of the earliest multimodal models developed specifically for medical applications, showing that few-shot multimodal prompting can be used for medical visual question answering~\cite{moor2023med}. Later work tested the same idea in more diagnostic classification settings. Ferber et al. showed that in-context examples improve GPT-4V performance on cancer histopathology tasks and can reach the level of specialist classifiers in some settings~\cite{ferber2024context}. Similar evidence has been reported in ophthalmology: Ayhan et al. reported that Gemini 1.5 Pro with ICL achieved diabetic retinopathy detection accuracy and F1 scores close to a supervised algorithm~\cite{ayhan2025context}, and Choi and Yoo demonstrated that example-based prompting improved GPT-4o performance across several ophthalmic image-interpretation tasks~\cite{choi2025evaluating}. In recent oncology work, Shrestha et al. compared several large multimodal language models under ICL across datasets such as MHIST, PatchCamelyon, and HAM10000, showing consistent gains from few-shot prompting while also indicating that fully fine-tuned systems still represent a strong ceiling~\cite{shrestha2025context}. These studies suggest that ICL can be useful in medical contexts when only a small number of labelled examples are available. However, the evidence is drawn primarily from pathology, oncology, and ophthalmology, and many studies rely on large externally hosted models.

Smaller model scales are particularly advantageous for wound classification. Wound images may be collected in hospitals, outpatient clinics, or home-care settings, and they often contain sensitive clinical information. In such cases, sending images to an external model provider may be challenging because of privacy constraints~\cite{zhong2025considerations}. Smaller models are therefore relevant not only because they are cheaper or faster, but because they can be run locally. Surveys of small and on-device language models describe these models as useful for lower-cost, lower-latency, and resource-constrained deployment~\cite{lu2024small,van2025survey,xu2024device}, and similar arguments have been made for small language models in healthcare~\cite{garg2025rise}. At the same time, local deployment is useful only if the performance loss is acceptable.

Few-shot prompting (or in-context learning) may benefit smaller models by providing task-specific information without fine-tuning or remote inference, but it raises another practical question: which examples should be shown to the model? In visual and multimodal ICL, performance can depend strongly on the selected examples, their order, and the retrieval strategy used to construct the prompt~\cite{chen2025can, li2024configure,luo2024does,zhang2023makes}. If the support examples are poorly matched to the query, the model may receive little useful task information, and if they are too similar to each other, the prompt may contain repeated evidence rather than a useful coverage of the class variation. A common solution is retrieval-based selection, where the support examples are chosen by nearest-neighbour search in an embedding space, often using a visual or vision-language encoder such as CLIP~\cite{radford2021clip}. Such retrieval-based strategies have been shown to improve over random example selection in visual ICL and are also used in multimodal ICL systems such as RICES~\cite{alayrac2022flamingo,luo2024does,zhang2023makes}. However, nearest-neighbour retrieval remains a limited criterion: it can select examples that are redundant, overly prototypical, or similar to the query for reasons irrelevant to the target class. This is especially important in fine-grained image tasks, where small visual differences may define the label. Recent work, therefore, explores support-selection strategies that go beyond simple similarity, including diversity-aware context search, learned retrievers, and counterfactual or contrastive example selection~\cite{kapuriya2025exploring,lee2026learning,suo2024visual,xiong2026retrieving}.

\section{Methods}

\subsection{Ethics statement}

This study does not include confidential information. All research procedures were conducted exclusively on publicly accessible, anonymised patient data, in compliance with all relevant ethical standards.

\subsection{Problem definition}

\begin{figure}[ht]
\centering
\includegraphics[width=\linewidth]{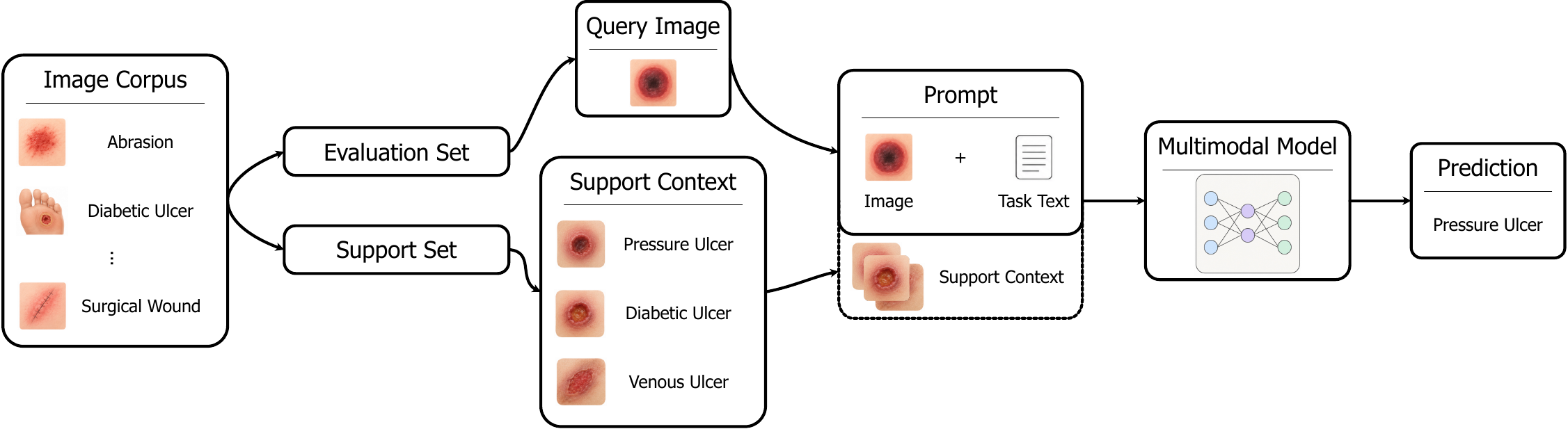}
\caption{Schematic overview of the ICL pipeline used in the current study. The wound images were generated with OpenAI's GPT Image 2 for visual illustration only.}
\label{fig:pipeline_schema}
\end{figure}

Let $\mathcal{D} = \{(I_j, y_j)\}_{j=1}^{N}$ denote an image corpus, where $I_j$ is an image and $y_j$ is its class label, belonging to a set of dataset labels $\mathcal{Y}$. To evaluate classification performance, $\mathcal{D}$ is divided into a support set $\mathcal{D}_{\mathrm{sup}}$ and an evaluation set $\mathcal{D}_{\mathrm{eval}}$. Given a query image $I_q \in \mathcal{D}_{\mathrm{eval}}$, we construct a prompt $\mathcal{P}_q$ containing the query image and task text. Optionally, the prompt also includes a retrieved support context $\mathcal{R} \subseteq \mathcal{D}_{\mathrm{sup}}$. The multimodal model $\Phi$ then predicts a class label $\hat{y}_q$ from this input. Inference can therefore be written as
\begin{equation*}
\hat{y}_q = \Phi(\mathcal{P}_q, \mathcal{R}).
\end{equation*}
In the zero-shot setting, $\mathcal{R} = \varnothing$, whereas in the ICL setting, $\mathcal{R}$ contains a set of retrieved, labelled support examples. Following prior work on multimodal in-context learning~\cite{ferber2024context, xiong2026retrieving}, we used embedding-based retrieval to construct this support context before passing it to the downstream classifier. Figure~\ref{fig:pipeline_schema} provides an overview of the full ICL pipeline.

\subsection{Datasets}

Experiments were performed on two wound‑image datasets. The first was the Kaggle dataset ``Collected and Categorized Wound Images Dataset''\footnote{Ibrahim Fateen, \textit{Collected and Categorized Wound Images Dataset}, Kaggle, \url{https://www.kaggle.com/datasets/ibrahimfateen/wound-classification} (accessed on 29 June 2026).}, which comprised 10 classes: \textit{Abrasions}, \textit{Bruises}, \textit{Burns}, \textit{Cut}, \textit{Diabetic Wounds}, \textit{Laceration}, \textit{Normal}, \textit{Pressure Wounds}, \textit{Surgical Wounds}, and \textit{Venous Wounds}. Before defining the final splits, half of the images were removed as they were mirrored duplicates of existing samples. The resulting Kaggle wound image corpus comprised 1469 images and was divided into a support split and an evaluation split. The evaluation split was balanced, containing 25 randomly selected images per class and 250 images in total, while the support split contained the remaining 1219 images.

The second dataset was constructed from the Medetec Wound Database\footnote{\raggedright Medetec, \textit{Medetec Wound Database: stock pictures of wounds}, \url{https://www.medetec.co.uk/files/medetec-image-databases.html} (accessed on 29 June 2026).}. After excluding low-count categories (less than 15 images), the Medetec corpus contained 560 images across 9 classes: \textit{Burns or Scalds}, \textit{Diabetic Foot Ulcers}, \textit{Extravasation Injuries}, \textit{Infected or Necrotic Toes}, \textit{Meningitis Skin Lesions}, \textit{Miscellaneous}, \textit{Orthopaedic Cases}, \textit{Pressure Ulcers}, and \textit{Venous or Arterial Ulcers}. As with the Kaggle dataset, the Medetec dataset was randomly divided into a support split and a balanced evaluation split. The evaluation set contained 10 images per class, for a total of 90 images; the remaining 470 images were used in the support set.

\begin{figure}[ht]
\centering
\includegraphics[width=\linewidth]{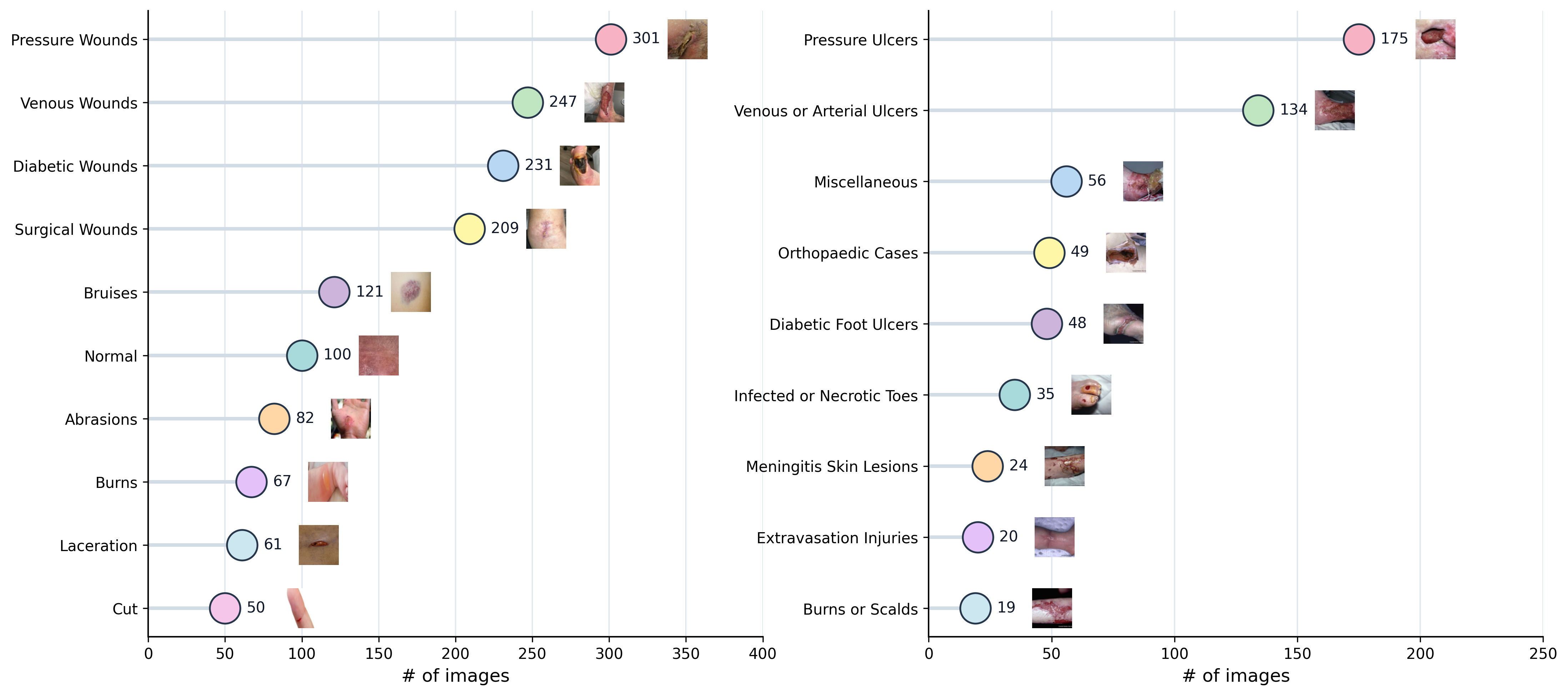}
\caption{Class distributions for the Kaggle dataset (left) and the Medetec dataset (right). Images are taken from the corresponding datasets.}
\label{fig:data_distribution_lollipop}
\end{figure}

For both datasets, the support splits served as the retrieval pools for few-shot context construction, and the evaluation splits were kept fixed for model assessment. Figure~\ref{fig:data_distribution_lollipop} shows the class distributions for the Kaggle dataset (left) and the Medetec dataset (right).

\subsection{Support retrieval}

For query-conditioned retrieval, the frozen OpenAI CLIP ViT-B/32 image encoder~\cite{radford2021clip} was used to map each image to a 512-dimensional embedding vector. For each support image $I_j \in \mathcal{D}_{\mathrm{sup}}$, an embedding was precomputed as
\begin{equation*}
z_{j} = f_I(I_j),
\end{equation*}
and for each query image $I_q \in \mathcal{D}_{\mathrm{eval}}$, an analogous embedding was computed as
\begin{equation*}
z_{q} = f_I(I_q).
\end{equation*}
To place all representations on a common scale, each embedding was L2-normalised,
\begin{equation*}
\hat{z}_i = \frac{z_i}{\lVert z_i \rVert_2},
\end{equation*}
so that angular proximity directly reflected semantic similarity in the shared feature space. Each candidate's support image was then scored using cosine similarity,
\begin{equation*}
s(q,i) = \hat{z}_q^{\top} \hat{z}_i.
\end{equation*}
This shared embedding space, therefore, served as the basis for all query-conditioned support selection. Three strategies were considered for constructing the few-shot support context: random retrieval, $k$-nearest-neighbour ($k$NN) retrieval (or Retrieval-based In-Context Example Selection, RICES~\cite{alayrac2022flamingo}), and $k$NN retrieval augmented with maximal marginal relevance (MMR)~\cite{carbonell1998use}.

\noindent\textbf{\textit{Random retrieval.}} Under random retrieval, support examples were sampled uniformly from $\mathcal{D}_{\mathrm{sup}}$ without conditioning on the query image, with a unique random seed used for each request.

\noindent\textbf{\textit{kNN retrieval.}} Given a query image $I_q$ and a target context size $K$, $k$NN retrieval constructs the support context by selecting the $K$ support images with the highest cosine-similarity scores $s(q,i)$. 

\noindent\textbf{\textit{kNN+MMR retrieval.}} To encourage diversity among relevant support examples, maximal marginal relevance was also used. In the $k$NN+MMR strategy, a candidate pool $\mathcal{C}_q$ of the top $M$ nearest neighbours of the query was retrieved according to $s(q, i)$, where $M > K$ and $K$ denotes the final context size. From $\mathcal{C}_q$, the highest-similarity item was chosen. Each subsequent example, $i_t$, was selected from the remaining candidates in $\mathcal{C}_q$ as %to balance query relevance against redundancy with the already selected context,
\begin{equation*}
i_t = \arg\max_{i \in \mathcal{C}_q \setminus \mathcal{R}_{t-1}} \left[ \lambda s(q,i) - (1-\lambda) \max_{j \in \mathcal{R}_{t-1}} \hat{z}_i^{\top} \hat{z}_j \right],
\end{equation*}
where $\lambda \in [0,1]$ controls the trade-off between relevance and diversity. In the current study, $\lambda$ was set to 0.7.

\subsection{Models}

We evaluated eleven multimodal language models from three families: Qwen 3.5 (0.8B, 2B, 4B, 9B, and 27B)~\cite{qwen}, Ministral 3 (3B, 8B, and 14B)~\cite{mistral2026ministral3}, and Gemma 4 (E2B, E4B, and 26B A4B)~\cite{gemma_2026_g4}. For consistency across experiments, all models were run with an 8192-token context window, thinking disabled, and temperature set to zero to make the outputs deterministic.

Experiments were run on a MacBook Pro with an Apple M4 Pro chip and 24 GB unified memory. The Qwen 3.5 checkpoints used \texttt{Q3\_K\_XL} quantisation. For Ministral 3, the 3B and 8B variants used \texttt{Q8\_0}, whereas the 14B variant used \texttt{Q4\_K\_XL}. For Gemma 4, the E2B and E4B variants used \texttt{Q8\_0}, whereas the 26B A4B variant used \texttt{UD-IQ3\_S}. Because all evaluated systems were multimodal, each model was paired with a vision projector; Gemma 4 and Ministral 3 used BF16 projectors, whereas Qwen 3.5 used an F32 projector.

\subsection{Prompts}

All models were queried using a classification prompt so that differences in performance could be attributed primarily to model capacity and support-selection strategy rather than to prompt wording. In every condition, the query image was paired with a short instruction that asked the model to assign exactly one dataset label and return only that label string. For the Kaggle dataset class labels were: \textit{Abrasions}, \textit{Bruises}, \textit{Burns}, \textit{Cut}, \textit{Diabetic Wounds}, \textit{Laceration}, \textit{Normal}, \textit{Pressure Wounds}, \textit{Surgical Wounds}, and \textit{Venous Wounds}; for the Medetec dataset: \textit{Burns or Scalds}, \textit{Diabetic Foot Ulcers}, \textit{Extravasation Injuries}, \textit{Infected or Necrotic Toes}, \textit{Meningitis Skin Lesions}, \textit{Miscellaneous}, \textit{Orthopaedic Cases}, \textit{Pressure Ulcers}, and \textit{Venous or Arterial Ulcers}.

In the zero-shot setting, the prompt consisted only of the task instruction and the query image, as illustrated in Figure~\ref{fig:zero_shot_imgref}. This setup measured each model's ability to classify wound images without any in-context visual references from the support set.

\begin{figure}[!b]
\centering
\fbox{%
\begin{minipage}{0.95\linewidth}
\small
\textbf{Zero-shot prompt}\par\medskip
\textbf{Query block:} \ImgRefPrompt\par
\textbf{Image:} query image $I_q$
\end{minipage}%
}
\caption{Zero-shot prompt structure.}
\label{fig:zero_shot_imgref}
\end{figure}

In the few-shot setting, the prompt began with a brief instruction informing the model that labelled example images would be shown before the final query. The prompt then included $K$ retrieved support examples, each presented as an image--label pair in the form ``\textit{Example image for class} $y_i$'', followed by the support image and its class label. After these support examples, the same query-classification block used in the zero-shot condition was appended, as shown in Figure~\ref{fig:few_shot_imgref}. For random retrieval and standard $k$NN retrieval, $K$ support examples were taken directly from the selected support set. For $k$NN+MMR retrieval, the final prompt still contained $K$ examples, but these were reranked from an initial pool of $M$ nearest neighbours before insertion into the prompt.

\begin{figure}[!h]
\centering
\fbox{%
\begin{minipage}{0.95\linewidth}
\small
\textbf{Few-shot prompt}\par\medskip
\textbf{Instruction block:} \textit{You will first see labelled example images for the dataset classes. Use them as visual references, then classify the final query image.}\par
\textbf{Support examples:} For each $i = 1, \dots, K$, the prompt includes the text ``\textit{Example image for class} $y_i$'', the corresponding support image $I_i$, and the class label $y_i$.\par
\textbf{Query block:} \ImgRefPrompt\par
\textbf{Image:} query image $I_q$
\end{minipage}%
}
\caption{Few-shot ($K$-shot) prompt structure.}
\label{fig:few_shot_imgref}
\end{figure}

\subsection{Retrieval influence diagnostics}

To assess how closely retrieval-augmented SMLM predictions aligned with the retrieved embedding-space neighbourhood during ICL~\cite{min2022rethinking, zhao2021calibrate}, paired comparisons were conducted between each model and a retrieval-only control. The same diagnostics were applied to both $k$NN and $k$NN+MMR retrieval. For standard $k$NN, the global weighted-$k$NN baseline was used as the control. For $k$NN+MMR, the retrieval-only control was recomputed on the final reranked support set, so that the control reflected the exact diversity-filtered context shown to the model.

The weighted-$k$NN baseline was treated as a standard non-parametric classifier operating directly in the CLIP embedding space. For a query $q$, let $\mathcal{N}_K(q)$ denote the $K$ labelled support images in $\mathcal{D}_{\mathrm{sup}}$ with the largest cosine similarities to the query. The class score was the weighted sum of neighbour labels, and the predicted class was
\begin{equation*}
\hat{y}_{q}^{\mathrm{kNN}} = \arg\max_{c \in \mathcal{Y}} \sum_{i \in \mathcal{N}_K(q)} w(q,i)\mathbf{1}[y_i=c],
\qquad
w(q,i)=\frac{\max\{s(q,i),0\}}{\sum_{j \in \mathcal{N}_K(q)}\max\{s(q,j),0\}},
\end{equation*}
Here, $c$ is a candidate class from the dataset-specific label set $\mathcal{Y}$, $y_i$ is neighbour $i$'s label, $\mathbf{1}[y_i=c]$ is the class-membership indicator, and $w(q,i)$ is the normalised non-negative cosine-similarity weight. For the matched $k$NN+MMR control, the same weighted-voting rule was applied to the final MMR-reranked neighbourhood shown in the prompt.

Statistical significance was assessed using accuracy, accuracy deltas, and a McNemar test, applied to correctness pairs to determine whether the SMLM significantly outperformed the baseline~\cite{mcnemar1947note}. To further characterise behavioural alignment, raw prediction agreement and Cohen's~$\kappa$ were computed, with the latter adjusting for agreement expected by chance~\cite{cohen1960coefficient}. Prediction alignment with the retrieved support labels was quantified using three match-rate diagnostics: the top-1 match rate, the majority-label match rate, and the any-support match rate. To distinguish cases in which the SMLM succeeded on the same items as the retrieval-only control from those in which it added value beyond retrieval, a paired $2\times2$ correctness table was also analysed for each model--control comparison. From this table, two conditional correctness statistics were computed: $P(\mathrm{SMLM\ correct}\mid\mathrm{control\ correct})$, which quantifies how often the SMLM preserves retrieval-only successes, and $P(\mathrm{SMLM\ correct}\mid\mathrm{control\ wrong})$, which quantifies how often the SMLM ``rescues'' retrieval-only failures. Fisher's exact test and the associated odds ratio were used to assess whether SMLM correctness was positively associated with control correctness beyond chance~\cite{fisher1970statistical}.

In these experiments, we used $K=10$ support images for $k$NN retrieval and $K=10$ with $M=30$ for $k$NN+MMR retrieval.

\subsection{Support set size impact}

To assess how the size of the support set $\mathcal{D}_{\mathrm{sup}}$ affects retrieval performance, we followed the procedure described by Xiong et al.~\cite{xiong2026retrieving}. In this procedure, the few-shot classification experiments are repeated for all retrieval methods using reduced support pools containing approximately 75\%~(914 for Kaggle, 353 for Medetec), 50\%~(610 for Kaggle, 235 for Medetec), and 25\%~(305 for Kaggle, 118 for Medetec) of the full support set. The full support set served as the benchmark. For each fraction, five subsets were sampled with different random seeds, while the evaluation set was kept fixed. To ensure fair comparisons, the same seed-specific subsets were used across all model and retrieval-method combinations within each fraction on both datasets. Performance on $\mathcal{D}_{\mathrm{eval}}$ was then summarised as mean $\pm$ standard deviation over the five runs.

Two descriptive statistics were computed for each sampled support subset to help assess whether changes in accuracy were more plausibly attributable to class imbalance or to reduced embedding-space diversity. Let $\{(I_i, y_i)\}_{i=1}^{n} \subseteq \mathcal{D}_{\mathrm{sup}}$ denote a sampled support subset of size $n$, and let $\mathcal{S} = \{(\hat{z}_i, y_i)\}_{i=1}^{n}$ be the corresponding set of embedding-label pairs, where $\hat{z}_i$ is the L2-normalized CLIP embedding of image $I_i$.

First, class-balance evenness was measured using the normalised Shannon entropy of the empirical class distribution, with larger values indicating a more even spread of support examples across $C$ classes. If $n_c$ is the number of support examples from class $c$ and $p_c = n_c / n$, then
\begin{equation*}
E_{\mathrm{class}} = - \frac{1}{\log C} \sum_{c=1}^{C} p_c \log p_c.
\end{equation*}

Second, the geometric diversity of the subset was measured through the effective rank of its embedding covariance spectrum. Let
\begin{equation*}
\Sigma_{\mathcal{S}} = \frac{1}{n} \sum_{i=1}^{n} (\hat{z}_i - \bar{z})(\hat{z}_i - \bar{z})^{\top}
\end{equation*}
be the covariance matrix of the subset embeddings, where $\bar{z} = \frac{1}{n}\sum_{i=1}^{n} \hat{z}_i$ is the sample mean. If $\{\sigma_m\}_{m=1}^{d}$ are the eigenvalues of $\Sigma_{\mathcal{S}}$, we normalize them as
\begin{equation*}
\tilde{\sigma}_m = \frac{\sigma_m}{\sum_{r=1}^{d} \sigma_r},
\end{equation*}
and define the effective rank as
\begin{equation*}
r_{\mathrm{eff}} = \exp\left(-\sum_{m=1}^{d} \tilde{\sigma}_m \log \tilde{\sigma}_m \right).
\end{equation*}
A larger effective rank indicates that variance is distributed across more dimensions, suggesting broader representational coverage of the support pool. For each support fraction, both $E_{\mathrm{class}}$ and $r_{\mathrm{eff}}$ were computed across the five seeds as mean $\pm$ standard deviation.

To evaluate how data scarcity affects ICL in comparison with conventional supervised methods, we additionally trained deep learning image classifiers on the same support sets used for ICL and repeated the experiment on subsets of both datasets. Each support pool was split internally into an 80/20 train/validation subset, and final performance was measured on the corresponding $\mathcal{D}_{\mathrm{eval}}$ sets. We evaluated four popular ImageNet-pretrained backbones: ViT-B/16~\cite{dosovitskiy2021image}, ConvNeXt-Tiny~\cite{liu2022convnet}, EfficientNetV2-S~\cite{tan2021efficientnetv2}, and ResNet50~\cite{he2016deep}. For each backbone, we compared full fine-tuning with a head-only transfer-learning setting in which the pretrained feature extractor was frozen and only the classification head was updated. For ViT-B/16, we additionally evaluated a LoRA variant ($r=8$, $\alpha=16$, dropout $0.05$) in which the pretrained ViT weights were frozen and only lightweight LoRA adapters plus the classification head were trained~\cite{hu2022lora}. All models were trained with AdamW optimiser, weight decay $10^{-4}$, batch size of 16 for 30 epochs with early stopping. Learning rates ranged from $1\times10^{-5}$ to $1\times10^{-3}$ depending on backbone and tuning mode. Training images were resized to the model input size and augmented with random horizontal flips ($p=0.5$), random rotations within $\pm8^\circ$, and colour jitter (brightness $0.12$, contrast $0.12$, saturation $0.08$), followed by ImageNet normalisation; validation and test images were only resized and normalised.

\subsection{Support context size impact}

To measure how context size affects both classification performance and inference cost, we conducted a support-count sweep in which the number of support images shown to the SMLM during ICL was varied while the support pool, evaluation set, prompt template, and model settings were kept fixed. We evaluated context sizes from $K=2$ to $K=20$ in steps of two images. For each condition, we report accuracy, the mean model-reported input token count, and the average response time per processed request.

\section{Results}

\subsection{In-context learning significantly improves classification performance}

Table~\ref{tab:benchmark_results} shows the results on the Kaggle wound dataset. Notably, few-shot prompting with query-conditioned retrieval consistently improved performance over zero-shot prompting by a substantial margin, whereas random few-shot context provided only modest gains. This indicates that the benefits from ICL depend largely on the retrieval strategy of support examples. The strongest overall results were obtained by Gemma~4 26B A4B and the larger Qwen 3.5 models: Gemma~4 26B A4B reached 0.868 accuracy under $k$NN (95\% CI [0.824, 0.908]), while Qwen~3.5 27B and 9B reached 0.872 and 0.868 under $k$NN+MMR (95\% CIs [0.828, 0.912] and [0.824, 0.908], respectively). By contrast, zero-shot performance was much lower and more variable, ranging from 0.128 for Qwen~3.5 0.8B (95\% CI [0.088, 0.172]) to 0.648 for Gemma~4 26B A4B (95\% CI [0.588, 0.704]).

The effect of query-specific retrieval relative to zero-shot and random few-shot prompting was especially pronounced for several smaller and mid-sized models. For example, Qwen~3.5 2B improved from 0.280 (95\% CI [0.228, 0.336]) in the zero-shot setting to 0.748 (95\% CI [0.696, 0.796]) with $k$NN+MMR, and Qwen~3.5 4B improved from 0.316 (95\% CI [0.260, 0.376]) to 0.788 (95\% CI [0.732, 0.836]). Similar gains were also observed for Ministral~3 14B, which increased from 0.364 (95\% CI [0.304, 0.424]) to 0.752 (95\% CI [0.692, 0.804]) under $k$NN, and for Gemma~4 E4B, which rose from 0.400 (95\% CI [0.336, 0.464]) to 0.736 (95\% CI [0.680, 0.788]). 
%Across model families, plain $k$NN was a strong and stable choice, and $k$NN+MMR provided small additional gains mainly within the Qwen family. 
Relative to the retrieval-only weighted-$k$NN baseline, which achieved 0.708 accuracy (95\% CI [0.648, 0.764]), the strongest query-conditioned ICL configurations achieved clear gains, whereas all zero-shot settings and several smaller retrieval-augmented models remained below this benchmark.

\begin{table*}[!ht]
\centering
\caption{Performance comparison of zero-shot and few-shot classification across models on the Kaggle dataset, reported in terms of accuracy and F1. The best result for each model is highlighted in bold. For both $k$NN and $k$NN+MMR, the final few-shot context contained $K=10$ support images; for $k$NN+MMR, these were selected from an initial candidate pool of $M=30$ nearest neighbors. The table also includes a retrieval-only, embedding-based weighted-$k$NN baseline for comparison.}
\label{tab:benchmark_results}
\small
\setlength{\tabcolsep}{4pt}
\begin{tabular}{l|cc|cc|cc|cc}
\toprule
\textbf{Model} & \multicolumn{2}{c|}{\makecell{\textbf{Zero-shot}}} & \multicolumn{2}{c|}{\makecell{\textbf{Few-shot} \\ \textbf{(Random)}}} & \multicolumn{2}{c|}{\makecell{\textbf{Few-shot} \\ \textbf{(kNN)}}} & \multicolumn{2}{c}{\makecell{\textbf{Few-shot} \\ \textbf{(kNN+MMR)}}} \\
& \textbf{Acc.} & \textbf{F1} & \textbf{Acc.} & \textbf{F1} & \textbf{Acc.} & \textbf{F1} & \textbf{Acc.} & \textbf{F1} \\
\midrule
\textbf{Qwen 3.5} & & & & & & & & \\
\quad 27B & 0.580 & 0.536 & 0.636 & 0.624 & 0.860 & 0.862 & \textbf{0.872} & \textbf{0.871} \\
\quad 9B & 0.544 & 0.510 & 0.640 & 0.615 & 0.836 & 0.833 & \textbf{0.868} & \textbf{0.865} \\
\quad 4B & 0.316 & 0.241 & 0.456 & 0.422 & 0.780 & 0.780 & \textbf{0.788} & \textbf{0.789} \\
\quad 2B & 0.280 & 0.173 & 0.324 & 0.266 & 0.704 & 0.680 & \textbf{0.748} & \textbf{0.726} \\
\quad 0.8B & 0.128 & 0.078 & 0.196 & 0.155 & \textbf{0.328} & \textbf{0.327} & 0.300 & 0.293 \\
\midrule
\textbf{Ministral 3} & & & & & & & & \\
\quad 14B & 0.364 & 0.348 & 0.448 & 0.447 & \textbf{0.752} & \textbf{0.758} & 0.708 & 0.712 \\
\quad 8B & 0.308 & 0.296 & 0.440 & 0.424 & \textbf{0.656} & 0.650 & \textbf{0.656} & \textbf{0.653} \\
\quad 3B & 0.308 & 0.251 & 0.392 & 0.334 & 0.548 & 0.525 & \textbf{0.580} & \textbf{0.558} \\
\midrule
\textbf{Gemma 4} & & & & & & & & \\
\quad 26B A4B & 0.648 & 0.643 & 0.652 & 0.647 & \textbf{0.868} & \textbf{0.868} & 0.864 & 0.864 \\
\quad E4B & 0.400 & 0.342 & 0.424 & 0.405 & \textbf{0.736} & \textbf{0.732} & 0.716 & 0.706 \\
\quad E2B & 0.304 & 0.262 & 0.332 & 0.296 & \textbf{0.456} & \textbf{0.434} & 0.444 & 0.426 \\
\midrule
\textit{Weighted-$k$NN} & \multicolumn{2}{c}{0.708~~~0.701} & \multicolumn{2}{c}{} & \multicolumn{2}{c}{} & \multicolumn{2}{c}{} \\
\bottomrule
\end{tabular}
\end{table*}

Table~\ref{tab:medetec_benchmark_results} shows the corresponding results on the Medetec dataset. The same broad pattern was observed, although absolute performance was lower than on the Kaggle dataset. The strongest overall result was obtained by Qwen~3.5 27B under $k$NN+MMR, reaching 0.678 accuracy. Other strong results were obtained by Qwen~3.5 4B, with 0.622 accuracy under $k$NN+MMR, Qwen~3.5 9B, with 0.611 accuracy under $k$NN+MMR, and Gemma~4 26B A4B, with 0.611 accuracy under $k$NN+MMR. By contrast, zero-shot performance was considerably lower, ranging from 0.089 for Qwen~3.5 0.8B (95\% CI [0.033, 0.156]) to 0.400 for Gemma~4 26B A4B (95\% CI [0.300, 0.500]). Random few-shot prompting produced only modest and inconsistent gains, and query-conditioned retrieval produced large gains across most models. Qwen~3.5 2B improved from chance-level 0.144 accuracy (95\% CI [0.078, 0.222]) in the zero-shot setting to 0.567 (95\% CI [0.467, 0.667]) with $k$NN+MMR, while Qwen~3.5 4B improved from 0.244 (95\% CI [0.156, 0.344]) to 0.622 (95\% CI [0.522, 0.722]). Similar improvements were observed for other model families: Ministral~3 14B increased from 0.244 accuracy (95\% CI [0.156, 0.333]) to 0.500 (95\% CI [0.400, 0.600]) under both $k$NN and $k$NN+MMR, and Gemma~4 E4B increased from 0.256 (95\% CI [0.167, 0.344]) to 0.533 (95\% CI [0.433, 0.644]) under $k$NN. As in the Kaggle dataset, the best retrieval-augmented multimodal configurations outperformed the retrieval-only weighted-$k$NN baseline, which reached 0.467 accuracy (95\% CI [0.367, 0.567]).

\begin{table*}[!t]
\centering
\caption{Performance comparison of zero-shot and few-shot classification across models on the Medetec dataset, reported in terms of accuracy and F1. The best result for each model is highlighted in bold. For both $k$NN and $k$NN+MMR, the final few-shot context contained $K=10$ support images; for $k$NN+MMR, these were selected from an initial candidate pool of $M=30$ nearest neighbors. The table also includes a retrieval-only, embedding-based weighted-$k$NN baseline for comparison.}
\label{tab:medetec_benchmark_results}
\small
\setlength{\tabcolsep}{4pt}
\begin{tabular}{l|cc|cc|cc|cc}
\toprule
\textbf{Model} & \multicolumn{2}{c|}{\makecell{\textbf{Zero-shot}}} & \multicolumn{2}{c|}{\makecell{\textbf{Few-shot} \\ \textbf{(Random)}}} & \multicolumn{2}{c|}{\makecell{\textbf{Few-shot} \\ \textbf{(kNN)}}} & \multicolumn{2}{c}{\makecell{\textbf{Few-shot} \\ \textbf{(kNN+MMR)}}} \\
& \textbf{Acc.} & \textbf{F1} & \textbf{Acc.} & \textbf{F1} & \textbf{Acc.} & \textbf{F1} & \textbf{Acc.} & \textbf{F1} \\
\midrule
\textbf{Qwen 3.5} & & & & & & & & \\
\quad 27B & 0.367 & 0.321 & 0.389 & 0.342 & 0.633 & 0.629 & \textbf{0.678} & \textbf{0.670} \\
\quad 9B  & 0.333 & 0.303 & 0.278 & 0.251 & 0.600 & 0.589 & \textbf{0.611} & \textbf{0.614} \\
\quad 4B  & 0.244 & 0.142 & 0.333 & 0.320 & 0.578 & 0.562 & \textbf{0.622} & \textbf{0.601} \\
\quad 2B  & 0.144 & 0.104 & 0.222 & 0.196 & 0.533 & 0.528 & \textbf{0.567} & \textbf{0.573} \\
\quad 0.8B & 0.089 & 0.031 & 0.167 & 0.112 & \textbf{0.244} & \textbf{0.220} & 0.200 & 0.176 \\
\midrule
\textbf{Ministral 3} & & & & & & & & \\
\quad 14B & 0.244 & 0.158 & 0.356 & 0.286 & \textbf{0.500} & \textbf{0.480} & \textbf{0.500} & 0.477 \\
\quad 8B  & 0.244 & 0.163 & 0.344 & 0.306 & 0.444 & 0.403 & \textbf{0.456} & \textbf{0.427} \\
\quad 3B  & 0.200 & 0.129 & 0.189 & 0.165 & 0.322 & 0.293 & \textbf{0.389} & \textbf{0.387} \\
\midrule
\textbf{Gemma 4} & & & & & & & & \\
\quad 26B A4B & 0.400 & 0.378 & 0.422 & 0.428 & 0.567 & 0.561 & \textbf{0.611} & \textbf{0.606} \\
\quad E4B     & 0.256 & 0.217 & 0.300 & 0.244 & \textbf{0.533} & \textbf{0.508} & 0.511 & 0.489 \\
\quad E2B     & 0.156 & 0.128 & 0.133 & 0.119 & \textbf{0.367} & \textbf{0.324} & 0.289 & 0.272 \\
\midrule
\textit{Weighted-$k$NN} & \multicolumn{2}{c}{0.467~~~0.447} & \multicolumn{2}{c}{} & \multicolumn{2}{c}{} & \multicolumn{2}{c}{} \\
\bottomrule
\end{tabular}
\end{table*}

Taken together, the Kaggle and Medetec results show a consistent hierarchy of prompting strategies. Zero-shot prompting was weakest, random few-shot prompting provided only limited and unreliable gains, and query-conditioned retrieval produced the largest improvements. Across model families, standard $k$NN was generally a strong and consistent retrieval strategy, whereas $k$NN+MMR usually provided only modest additional gains, primarily within the Qwen family. The Kaggle dataset yielded higher absolute performance, with top accuracy above 0.87, whereas the best Medetec result reached 0.678 accuracy. Nevertheless, the main trend was stable across both datasets: retrieval-based ICL accounted for most of the performance lift, while the stronger multimodal SMLMs surpassed the retrieval-only baseline by a clear margin.

\subsection{Larger models are more likely to improve beyond retrieval-only baselines}
\label{sec:results_retrieval_control}

For the Kaggle dataset, Table~\ref{tab:retrieval_control_knn} summarises the primary paired comparison against the retrieval-only control for $k$NN ICL: Qwen~3.5 27B, Qwen~3.5 9B, and Gemma~4 26B A4B exceeded the weighted-$k$NN baseline (0.708 accuracy) by +0.152, +0.128, and +0.160, respectively, with exact McNemar tests indicating significant paired improvements. At the same time, their support-overlap rates remained high: top-1 support matches ranged from 0.708 to 0.732, support-majority matches from 0.720 to 0.744, and any-support matches from 0.952 to 0.980. Conditional correctness analysis showed the same pattern: for these three models, the probability of being correct when the weighted-$k$NN control was correct remained very high (0.938--0.966), while the probability of being correct when the weighted-$k$NN control was wrong was still substantial (0.589--0.630), with Fisher exact tests indicating strong positive association between SMLM and control correctness (odds ratios $=10.53$--$16.73$, all $p<10^{-9}$; see Appendix~\ref{sec:supp_retrieval_control_association}). This pattern indicates that their predictions were strongly constrained by the retrieved support, yet were not reducible to a simple nearest-neighbour vote, because agreement and $\kappa$ remained substantial rather than near-perfect, and the models still achieved significant gains over the weighted-$k$NN control. Qwen~3.5 4B also remained above the weighted-$k$NN control (+0.072; McNemar $p=0.011$), but by a smaller margin; correspondingly, it preserved most weighted-$k$NN-correct cases (0.921) but rescued a more modest fraction of weighted-$k$NN errors (0.438). 

\begin{table*}[!t]
\centering
\caption{Paired retrieval-control comparisons for $k$NN few-shot prompting against the global weighted-$k$NN control (accuracy = 0.708) on the Kaggle dataset. Top-1, Maj., and Any denote match rates to the top retrieved support label, the support-majority label, and any label present in the retrieved support set, respectively. Ret.\ denotes $P(\mathrm{SMLM\ correct}\mid\mathrm{control\ correct})$, and Resc.\ denotes $P(\mathrm{SMLM\ correct}\mid\mathrm{control\ wrong})$. Agreement denotes the raw prediction agreement rate with the control, and $\kappa$ denotes Cohen's kappa.}
\label{tab:retrieval_control_knn}
\scriptsize
\setlength{\tabcolsep}{3pt}
\begin{tabular}{lcccccccccc}
\toprule
\textbf{Model} & \textbf{Acc.} & \textbf{$\Delta$} & \textbf{McNemar $p$} & \textbf{Top-1} & \textbf{Maj.} & \textbf{Any} & \textbf{Ret.} & \textbf{Resc.} & \textbf{Agreement} & \textbf{$\kappa$} \\
\midrule
Qwen~3.5 27B     & 0.860 & +0.152 & $<0.001$ & 0.732 & 0.744 & 0.980 & 0.960 & 0.616 & 0.752 & 0.723 \\
Qwen~3.5 9B      & 0.836 & +0.128 & $<0.001$ & 0.716 & 0.740 & 0.964 & 0.938 & 0.589 & 0.740 & 0.710 \\
Gemma~4 26B A4B  & 0.868 & +0.160 & $<0.001$ & 0.708 & 0.720 & 0.952 & 0.966 & 0.630 & 0.728 & 0.697 \\
Qwen~3.5 4B      & 0.780 & +0.072 & 0.011    & 0.704 & 0.752 & 0.984 & 0.921 & 0.438 & 0.744 & 0.713 \\
Ministral~3 14B  & 0.752 & +0.044 & 0.169    & 0.684 & 0.680 & 0.944 & 0.881 & 0.438 & 0.692 & 0.656 \\
Gemma~4 E4B      & 0.736 & +0.028 & 0.401    & 0.720 & 0.716 & 0.948 & 0.876 & 0.397 & 0.712 & 0.676 \\
Qwen~3.5 2B      & 0.704 & -0.004 & 1.000    & 0.676 & 0.716 & 0.928 & 0.853 & 0.342 & 0.720 & 0.687 \\
Ministral~3 8B   & 0.656 & -0.052 & 0.130    & 0.640 & 0.632 & 0.884 & 0.785 & 0.342 & 0.640 & 0.595 \\
Ministral~3 3B   & 0.548 & -0.160 & $<0.001$ & 0.528 & 0.508 & 0.768 & 0.633 & 0.342 & 0.504 & 0.449 \\
Gemma~4 E2B      & 0.456 & -0.252 & $<0.001$ & 0.436 & 0.432 & 0.668 & 0.520 & 0.301 & 0.424 & 0.358 \\
Qwen~3.5 0.8B    & 0.328 & -0.380 & $<0.001$ & 0.296 & 0.292 & 0.608 & 0.362 & 0.247 & 0.296 & 0.207 \\
\bottomrule
\end{tabular}
\end{table*}

By contrast, the weaker models showed lower agreement with the weighted-$k$NN control and weaker support overlap. Qwen~3.5 2B was already at parity with the control ($\Delta=-0.004$), although still demonstrating strong conditional dependence on the weighted-$k$NN baseline (retention 0.853, rescue 0.342; Fisher $p<10^{-14}$), the cases it rescued were nearly offset by cases in which the control had been correct, and the SMLM was not. Smaller models such as Gemma~4 E2B, Ministral~3 3B, and Qwen~3.5 0.8B had substantially lower agreement values (0.424, 0.504, and 0.296, respectively) and $\kappa$ values (0.358, 0.449, and 0.207), with lower any-support overlap (0.668, 0.768, and 0.608). Together with their lower conditional retention and rescue rates, this suggests that the smaller models were less consistently anchored to the retrieved neighbourhood and made less effective use of ICL.

For the Medetec dataset, Table~\ref{tab:medetec_retrieval_control_knn} summarises the analogous paired retrieval-control comparisons for the $k$NN retrieval. The global weighted-$k$NN control reached 0.467 accuracy, and Qwen~3.5 27B and 9B exceeded it by +0.167 and +0.133, with an exact McNemar test indicating a significant paired improvement. By contrast, Gemma~4 26B A4B and Qwen~3.5 4B achieved positive deltas of +0.100 and +0.111, respectively, but these differences were not significant. The strongest models remained closely tied to the retrieved support: Qwen~3.5 27B had a support-majority match rate of 0.744 and an any-support match rate of 0.978, while Qwen~3.5 9B had an any-support match rate of 1.000. Conditional correctness analysis showed that these models preserved most control-correct cases, with retention rates of 0.952 and 0.857, while rescuing a smaller fraction of control errors, with rescue rates of 0.354 and 0.375. Fisher's exact tests indicated a strong positive association between SMLM and control correctness for both models (odds ratios $=36.47$ and $10.00$; see Appendix~\ref{sec:supp_retrieval_control_association}).

Several models showed a different pattern on Medetec: they were strongly coupled to the retrieval-only control but did not rescue enough control errors to obtain a significant net improvement. Gemma~4 E4B, for example, had high retention (0.905) and a strong association with the control (odds ratio $=36.10$), but its rescue rate was only 0.208, and its accuracy gain over the control was not significant. Similarly, Qwen~3.5 2B had a positive but non-significant delta of +0.067. At the lower end, Ministral~3 3B and Qwen~3.5 0.8B fell significantly below the retrieval-only control. Qwen~3.5 0.8B was particularly weakly anchored to the retrieved neighbourhood, with low support-overlap values, very low agreement with the control, and no significant association between SMLM and control correctness.

\begin{table*}[!t]
\centering
\caption{Paired retrieval-control comparisons for $k$NN few-shot prompting against the global weighted-$k$NN control (accuracy = 0.467) on the Medetec dataset. Top-1, Maj., and Any denote match rates to the top retrieved support label, the support-majority label, and any label present in the retrieved support set, respectively. Ret.\ denotes $P(\mathrm{SMLM\ correct}\mid\mathrm{control\ correct})$, and Resc.\ denotes $P(\mathrm{SMLM\ correct}\mid\mathrm{control\ wrong})$. Agreement denotes the raw prediction agreement rate with the control, and $\kappa$ denotes Cohen's kappa.}
\label{tab:medetec_retrieval_control_knn}
\scriptsize
\setlength{\tabcolsep}{3pt}
\begin{tabular}{lcccccccccc}
\toprule
\textbf{Model} & \textbf{Acc.} & \textbf{$\Delta$} & \textbf{McNemar $p$} & \textbf{Top-1} & \textbf{Maj.} & \textbf{Any} & \textbf{Ret.} & \textbf{Resc.} & \textbf{Agreement} & \textbf{$\kappa$} \\
\midrule
Qwen~3.5 27B        & 0.633 & +0.167 & $<0.001$ & 0.611 & 0.744 & 0.978 & 0.952 & 0.354 & 0.733 & 0.660 \\
Qwen~3.5 9B         & 0.600 & +0.133 & 0.023    & 0.522 & 0.589 & 1.000 & 0.857 & 0.375 & 0.578 & 0.498 \\
Gemma~4 26B A4B     & 0.567 & +0.100 & 0.108    & 0.522 & 0.544 & 0.911 & 0.810 & 0.354 & 0.533 & 0.446 \\
Qwen~3.5 4B         & 0.578 & +0.111 & 0.076    & 0.433 & 0.567 & 0.844 & 0.810 & 0.375 & 0.556 & 0.483 \\
Ministral~3 14B     & 0.500 & +0.033 & 0.701    & 0.533 & 0.544 & 0.933 & 0.714 & 0.312 & 0.533 & 0.441 \\
Gemma~4 E4B         & 0.533 & +0.067 & 0.180    & 0.600 & 0.700 & 0.978 & 0.905 & 0.208 & 0.689 & 0.608 \\
Qwen~3.5 2B         & 0.533 & +0.067 & 0.362    & 0.444 & 0.444 & 0.922 & 0.714 & 0.375 & 0.433 & 0.364 \\
Ministral~3 8B      & 0.444 & -0.022 & 0.839    & 0.500 & 0.511 & 0.811 & 0.690 & 0.229 & 0.511 & 0.411 \\
Gemma~4 E2B         & 0.367 & -0.100 & 0.150    & 0.422 & 0.289 & 0.744 & 0.524 & 0.229 & 0.289 & 0.216 \\
Ministral~3 3B      & 0.322 & -0.144 & 0.029    & 0.289 & 0.322 & 0.767 & 0.476 & 0.188 & 0.311 & 0.240 \\
Qwen~3.5 0.8B       & 0.244 & -0.222 & 0.005    & 0.156 & 0.100 & 0.467 & 0.214 & 0.271 & 0.100 & 0.041 \\
\bottomrule
\end{tabular}
\end{table*}

Because the $k$NN+MMR retrieval changed only the final support-set selection step, its retrieval-control diagnostics are not repeated in full here. Across both datasets, the $k$NN+MMR results followed the same qualitative pattern as standard $k$NN: larger models retained most retrieval-control successes while rescuing a subset of retrieval-control failures, whereas smaller models showed weaker evidence of reasoning beyond the retrieved neighbourhood. Detailed $k$NN+MMR paired results are reported in Appendix~\ref{sec:supp_mmr_retrieval_control}.

Overall, the results show that larger models make more effective use of ICL without merely ``parroting'' the support context. Conditional correctness analysis makes the size distinction clearer: the larger models preserve most cases that the retrieval-only control gets right while also ``rescuing'' a subset of the control's errors. In other words, they remain guided by the retrieved neighbourhood while still outperforming retrieval-only controls on the same examples, and their agreement with those controls is substantial but far from perfect. The smaller models also benefit from relevant retrieved examples compared with zero-shot and random few-shot prompting, but they either use ICL without achieving any improvement or fail to exploit it reliably, consistently falling below what retrieval alone achieves on the same context.

\subsection{ICL performance with SMLMs degrades modestly under reduced support}

Table~\ref{tab:subset_stats} shows that the reduced support subsets remained broadly class-balanced across fractions. For Kaggle, class-balance evenness changed only slightly, from 0.889 for the full support set to 0.879 at the 25\% fraction. Medetec showed a similarly small mean change, from 0.808 to 0.799, although variability increased at the smallest support size. By contrast, the effective rank of the CLIP embedding space decreased more clearly as support size was reduced, from 52.9 to 45.5 for Kaggle and from 56.0 to 38.3 for Medetec. This suggests that the support-size ablation primarily reduced embedding-space coverage rather than causing large class-balance shifts, particularly for Medetec where the smaller support pool lost representational diversity more quickly.

\begin{table}[!t]
\centering
\caption{Support-set size and subset statistics across the support-set-size ablation for different subset selection seeds: subset size ($n$), class-balance evenness ($E_{\mathrm{class}}$), and effective rank ($r_{\mathrm{eff}}$).}
\label{tab:subset_stats}
\small
\setlength{\tabcolsep}{6pt}
\begin{tabular}{llcccc}
\toprule
\textbf{Dataset} & \textbf{Metric} & \textbf{100\%} & \textbf{75\%} & \textbf{50\%} & \textbf{25\%} \\
\midrule
\multirow{3}{*}{Kaggle} & $n$ & 1219 & 914 & 610 & 305 \\
 & $E_{\mathrm{class}}$ & 0.889 & 0.889 $\pm$ 0.002 & 0.888 $\pm$ 0.006 & 0.879 $\pm$ 0.005 \\
 & $r_{\mathrm{eff}}$ & 52.9 & 51.7 $\pm$ 0.1 & 49.3 $\pm$ 0.6 & 45.5 $\pm$ 1.2 \\
\midrule
\multirow{3}{*}{Medetec} & $n$ & 470 & 353 & 235 & 118 \\
 & $E_{\mathrm{class}}$ & 0.808 & 0.809 $\pm$ 0.009 & 0.804 $\pm$ 0.012 & 0.799 $\pm$ 0.048 \\
 & $r_{\mathrm{eff}}$ & 56.0 & 53.2 $\pm$ 0.5 & 48.1 $\pm$ 0.4 & 38.3 $\pm$ 0.8 \\
\bottomrule
\end{tabular}
\end{table}

Table~\ref{tab:mixed_accuracy_support_sweep} reports the accuracy trends across support fractions on the Kaggle dataset. The strongest retrieval-based SMLMs degraded only moderately as the support pool was reduced. Qwen~3.5 27B with $k$NN+MMR changed from 0.872 at full support to 0.830 at the 50\% fraction ($-4.8\%$) and 0.781 at the 25\% fraction ($-10.4\%$). Gemma~4 26B A4B was also robust, reaching 0.829 at the 50\% fraction ($-4.5\%$) and 0.794 at the 25\% fraction ($-8.5\%$) with $k$NN+MMR, and 0.848 at the 50\% fraction ($-1.9\%$) and 0.798 at the 25\% fraction ($-7.6\%$) with standard $k$NN. Qwen~3.5 27B with standard $k$NN and Qwen~3.5 9B with $k$NN+MMR showed similar reductions at the 25\% fraction, reaching 0.769 ($-10.6\%$) and 0.763 ($-12.1\%$), respectively. The supervised baselines lost more performance at the smallest training size: the best supervised result at 25\% was 0.666 for ConvNeXt-Tiny with full fine-tuning, down from 0.804 at full support ($-17.2\%$). 
%Thus, on Kaggle, retrieval-based prompting preserved most of its full-support performance and remained competitive with, or above, supervised fine-tuning across all support fractions.

The lower-capacity and random-context conditions showed a different pattern. Random few-shot prompting remained substantially below query-conditioned retrieval throughout the sweep, and its accuracy changed only weakly with support-pool size, suggesting that the main advantage came from retrieving relevant examples rather than from simply having a larger pool of support images. Smaller retrieval-augmented models also benefited from $k$NN or $k$NN+MMR selection, but their absolute performance was lower, and their trends were less monotonic. This is consistent with the earlier retrieval-control results: retrieved context remained useful, but the strongest gains under support reduction were concentrated in the larger SMLMs.

Table~\ref{tab:medetec_mixed_accuracy_support_sweep} shows the same support-fraction analysis for the Medetec dataset. Absolute accuracies were lower than on Kaggle, and the reduction from full support was larger for several methods. Qwen~3.5 27B with $k$NN+MMR decreased from 0.678 to 0.527 at the 25\% fraction ($-22.3\%$), while the $k$NN version decreased from 0.633 to 0.518 ($-18.2\%$). Gemma~4 26B A4B with $k$NN+MMR was more stable, decreasing from 0.611 to 0.529 ($-13.4\%$). These retrieval-based SMLM results still exceeded the supervised baselines at the smallest training size. At the 25\% fraction, the best SMLM results were around 0.52--0.53 accuracy, whereas the best supervised result was 0.413 for ConvNeXt-Tiny with head fine-tuning, down from 0.644 at full support ($-35.9\%$).

The stronger degradation on Medetec is consistent with the subset statistics in Table~\ref{tab:subset_stats}. The effective rank decreased from 56.0 to 38.3 on Medetec ($-31.6\%$), compared with 52.9 to 45.5 on Kaggle ($-14.0\%$). Thus, the smaller Medetec support pool lost embedding-space coverage more quickly. Nevertheless, the strongest retrieval-based SMLM conditions retained a clear advantage over both random prompting and supervised training at the smallest fractions. Overall, Tables~\ref{tab:mixed_accuracy_support_sweep} and~\ref{tab:medetec_mixed_accuracy_support_sweep} indicate that retrieval-based ICL is fairly robust to moderate support reduction, but that performance starts to decline when the support pool becomes too small or too narrow.

\begin{table}[H]
\caption{Accuracy across support fractions on the Kaggle dataset. The 1.0 column contains single-run values; the other columns show mean $\pm$ std across five seeds. Rows are grouped into supervised image models and non-supervised retrieval-based methods. For supervised models, \textit{full} denotes full fine-tuning with an unfrozen backbone, \textit{LoRA} denotes training lightweight low-rank adapters, and \textit{head} denotes training only the final classification layer with the backbone frozen.}
\centering
\small
\setlength{\tabcolsep}{6pt}
\begin{tabular}{lcccc}
\toprule
Model & 1.0 & 0.75 & 0.50 & 0.25 \\
\midrule
\multicolumn{5}{l}{\textbf{Supervised models}} \\
\midrule
ViT-B/16 full & 0.808 & 0.766 $\pm$ 0.027 & 0.706 $\pm$ 0.023 & 0.642 $\pm$ 0.019 \\
ConvNeXt-Tiny full & 0.804 & 0.779 $\pm$ 0.022 & 0.765 $\pm$ 0.025 & 0.666 $\pm$ 0.028 \\
ViT-B/16 LoRA r=8 & 0.788 & 0.778 $\pm$ 0.025 & 0.733 $\pm$ 0.023 & 0.648 $\pm$ 0.037 \\
EfficientNet V2-S full & 0.772 & 0.786 $\pm$ 0.014 & 0.716 $\pm$ 0.024 & 0.626 $\pm$ 0.061 \\
ResNet50 full & 0.760 & 0.734 $\pm$ 0.028 & 0.685 $\pm$ 0.013 & 0.626 $\pm$ 0.044 \\
ViT-B/16 head & 0.716 & 0.718 $\pm$ 0.036 & 0.683 $\pm$ 0.030 & 0.639 $\pm$ 0.026 \\
ConvNeXt-Tiny head & 0.700 & 0.730 $\pm$ 0.025 & 0.698 $\pm$ 0.010 & 0.643 $\pm$ 0.027 \\
ResNet50 head & 0.676 & 0.656 $\pm$ 0.041 & 0.603 $\pm$ 0.031 & 0.560 $\pm$ 0.022 \\
EfficientNet V2-S head & 0.644 & 0.562 $\pm$ 0.024 & 0.519 $\pm$ 0.026 & 0.482 $\pm$ 0.015 \\
\midrule
\multicolumn{5}{l}{\textbf{SMLMs}} \\
\midrule
Qwen 3.5 27B kNN+MMR & 0.872 & 0.873 $\pm$ 0.011 & 0.830 $\pm$ 0.040 & 0.781 $\pm$ 0.025 \\
Gemma 4 26B A4B kNN+MMR & 0.868 & 0.852 $\pm$ 0.016 & 0.829 $\pm$ 0.019 & 0.794 $\pm$ 0.019 \\
Qwen 3.5 9B kNN+MMR & 0.868 & 0.849 $\pm$ 0.010 & 0.823 $\pm$ 0.017 & 0.763 $\pm$ 0.010 \\
Gemma 4 26B A4B kNN & 0.864 & 0.845 $\pm$ 0.006 & 0.848 $\pm$ 0.025 & 0.798 $\pm$ 0.022 \\
Qwen 3.5 27B kNN & 0.860 & 0.851 $\pm$ 0.013 & 0.829 $\pm$ 0.012 & 0.769 $\pm$ 0.013 \\
Qwen 3.5 9B kNN & 0.836 & 0.825 $\pm$ 0.017 & 0.816 $\pm$ 0.019 & 0.743 $\pm$ 0.022 \\
Ministral 3 14B kNN & 0.752 & 0.694 $\pm$ 0.016 & 0.671 $\pm$ 0.010 & 0.628 $\pm$ 0.026 \\
Qwen 3.5 2B kNN+MMR & 0.748 & 0.719 $\pm$ 0.018 & 0.705 $\pm$ 0.013 & 0.646 $\pm$ 0.020 \\
Gemma 4 E4B kNN & 0.736 & 0.706 $\pm$ 0.011 & 0.680 $\pm$ 0.014 & 0.623 $\pm$ 0.007 \\
Gemma 4 E4B kNN+MMR & 0.716 & 0.692 $\pm$ 0.012 & 0.696 $\pm$ 0.006 & 0.647 $\pm$ 0.017 \\
Ministral 3 14B kNN+MMR & 0.708 & 0.710 $\pm$ 0.020 & 0.674 $\pm$ 0.029 & 0.633 $\pm$ 0.016 \\
Qwen 3.5 2B kNN & 0.704 & 0.713 $\pm$ 0.009 & 0.694 $\pm$ 0.025 & 0.637 $\pm$ 0.018 \\
Ministral 3 8B kNN+MMR & 0.656 & 0.639 $\pm$ 0.008 & 0.611 $\pm$ 0.016 & 0.573 $\pm$ 0.013 \\
Ministral 3 8B kNN & 0.656 & 0.630 $\pm$ 0.017 & 0.602 $\pm$ 0.026 & 0.573 $\pm$ 0.026 \\
Gemma 4 26B A4B Random & 0.652 & 0.640 $\pm$ 0.016 & 0.644 $\pm$ 0.005 & 0.642 $\pm$ 0.023\\
Qwen 3.5 9B Random & 0.640 & 0.619 $\pm$ 0.020 & 0.616 $\pm$ 0.014 & 0.624 $\pm$ 0.017 \\
Qwen 3.5 27B Random & 0.636 & 0.623 $\pm$ 0.009 & 0.609 $\pm$ 0.014 & 0.607 $\pm$ 0.025 \\
Ministral 3 3B kNN+MMR & 0.580 & 0.563 $\pm$ 0.017 & 0.553 $\pm$ 0.023 & 0.506 $\pm$ 0.029 \\
Ministral 3 3B kNN & 0.548 & 0.534 $\pm$ 0.016 & 0.529 $\pm$ 0.026 & 0.496 $\pm$ 0.019 \\
Gemma 4 E2B kNN & 0.456 & 0.494 $\pm$ 0.012 & 0.508 $\pm$ 0.020 & 0.476 $\pm$ 0.007 \\
Ministral 3 14B Random & 0.448 & 0.449 $\pm$ 0.014 & 0.459 $\pm$ 0.027 & 0.457 $\pm$ 0.030 \\
Gemma 4 E2B kNN+MMR & 0.444 & 0.469 $\pm$ 0.013 & 0.483 $\pm$ 0.010 & 0.461 $\pm$ 0.014 \\
Gemma 4 E4B Random & 0.424 & 0.452 $\pm$ 0.006 & 0.444 $\pm$ 0.008 & 0.445 $\pm$ 0.007 \\
Ministral 3 3B Random & 0.392 & 0.400 $\pm$ 0.013 & 0.410 $\pm$ 0.011 & 0.419 $\pm$ 0.013 \\
Ministral 3 8B Random & 0.440 & 0.383 $\pm$ 0.013 & 0.403 $\pm$ 0.023 & 0.417 $\pm$ 0.028 \\
Gemma 4 E2B Random & 0.332 & 0.309 $\pm$ 0.027 & 0.339 $\pm$ 0.024 & 0.331 $\pm$ 0.017 \\
Qwen 3.5 2B Random & 0.324 & 0.351 $\pm$ 0.014 & 0.356 $\pm$ 0.013 & 0.341 $\pm$ 0.018 \\
\bottomrule
\end{tabular}
\label{tab:mixed_accuracy_support_sweep}
\end{table}

\begin{table}[!t]
\caption{Accuracy across support fractions on the Medetec dataset. The 1.0 column contains single-run values; the other columns show mean $\pm$ std across five seeds. Rows are grouped into supervised image models and non-supervised retrieval-based methods. For supervised models, \textit{full} denotes full fine-tuning with an unfrozen backbone, \textit{LoRA} denotes training lightweight low-rank adapters, and \textit{head} denotes training only the final classification layer with the backbone frozen.}
\centering
\small
\setlength{\tabcolsep}{6pt}
\begin{tabular}{lcccc}
\toprule
Model & 1.0 & 0.75 & 0.50 & 0.25 \\
\midrule
\multicolumn{5}{l}{\textbf{Supervised image models}} \\
\midrule
ConvNeXt-Tiny head & 0.644 & 0.578 $\pm$ 0.008 & 0.522 $\pm$ 0.055 & 0.413 $\pm$ 0.096 \\
ConvNeXt-Tiny full & 0.567 & 0.551 $\pm$ 0.033 & 0.491 $\pm$ 0.053 & 0.384 $\pm$ 0.061 \\
ViT-B/16 head & 0.567 & 0.511 $\pm$ 0.054 & 0.471 $\pm$ 0.068 & 0.398 $\pm$ 0.085 \\
ViT-B/16 LoRA r=8 & 0.556 & 0.511 $\pm$ 0.053 & 0.480 $\pm$ 0.064 & 0.389 $\pm$ 0.067 \\
EfficientNet V2-S full & 0.556 & 0.487 $\pm$ 0.055 & 0.467 $\pm$ 0.072 & 0.327 $\pm$ 0.081 \\
ResNet50 head & 0.556 & 0.507 $\pm$ 0.026 & 0.442 $\pm$ 0.043 & 0.287 $\pm$ 0.100 \\
ViT-B/16 full & 0.533 & 0.529 $\pm$ 0.006 & 0.462 $\pm$ 0.069 & 0.376 $\pm$ 0.062 \\
ResNet50 full & 0.522 & 0.489 $\pm$ 0.050 & 0.427 $\pm$ 0.052 & 0.353 $\pm$ 0.084 \\
EfficientNet V2-S head & 0.433 & 0.422 $\pm$ 0.021 & 0.396 $\pm$ 0.051 & 0.293 $\pm$ 0.045 \\
\midrule
\multicolumn{5}{l}{\textbf{SMLMs}} \\
\midrule
Qwen 3.5 27B kNN+MMR & 0.678 & 0.633 $\pm$ 0.023 & 0.577 $\pm$ 0.022 & 0.527 $\pm$ 0.074 \\
Qwen 3.5 27B kNN & 0.633 & 0.611 $\pm$ 0.021 & 0.555 $\pm$ 0.013 & 0.518 $\pm$ 0.076 \\
Gemma 4 26B A4B kNN+MMR & 0.611 & 0.604 $\pm$ 0.006 & 0.582 $\pm$ 0.056 & 0.529 $\pm$ 0.077 \\
Qwen 3.5 9B kNN+MMR & 0.611 & 0.562 $\pm$ 0.036 & 0.531 $\pm$ 0.021 & 0.458 $\pm$ 0.068 \\
Qwen 3.5 9B kNN & 0.600 & 0.578 $\pm$ 0.021 & 0.513 $\pm$ 0.034 & 0.467 $\pm$ 0.053 \\
Gemma 4 26B A4B kNN & 0.567 & 0.569 $\pm$ 0.018 & 0.540 $\pm$ 0.027 & 0.522 $\pm$ 0.060\\
Qwen 3.5 2B kNN+MMR & 0.567 & 0.529 $\pm$ 0.023 & 0.467 $\pm$ 0.057 & 0.404 $\pm$ 0.076 \\
Gemma 4 E4B kNN & 0.533 & 0.524 $\pm$ 0.033 & 0.449 $\pm$ 0.034 & 0.427 $\pm$ 0.059 \\
Qwen 3.5 2B kNN & 0.533 & 0.513 $\pm$ 0.009 & 0.433 $\pm$ 0.050 & 0.382 $\pm$ 0.048 \\
Gemma 4 E4B kNN+MMR & 0.511 & 0.493 $\pm$ 0.029 & 0.469 $\pm$ 0.043 & 0.431 $\pm$ 0.033 \\
Ministral 3 14B kNN+MMR & 0.500 & 0.482 $\pm$ 0.025 & 0.440 $\pm$ 0.006 & 0.422 $\pm$ 0.048 \\
Ministral 3 14B kNN & 0.500 & 0.449 $\pm$ 0.025 & 0.443 $\pm$ 0.008 & 0.409 $\pm$ 0.043 \\
Ministral 3 8B kNN+MMR & 0.456 & 0.436 $\pm$ 0.024 & 0.407 $\pm$ 0.023 & 0.389 $\pm$ 0.038 \\
Ministral 3 8B kNN & 0.444 & 0.442 $\pm$ 0.016 & 0.424 $\pm$ 0.042 & 0.389 $\pm$ 0.046 \\
Gemma 4 26B A4B Random & 0.422 & 0.415 $\pm$ 0.050 & 0.427 $\pm$ 0.033 & 0.416 $\pm$ 0.042 \\
Ministral 3 3B kNN+MMR & 0.389 & 0.327 $\pm$ 0.017 & 0.353 $\pm$ 0.041 & 0.309 $\pm$ 0.032 \\
Qwen 3.5 27B Random & 0.388 & 0.402 $\pm$ 0.015 & 0.406 $\pm$ 0.020 & 0.424 $\pm$ 0.033 \\
Gemma 4 E2B kNN & 0.367 & 0.296 $\pm$ 0.030 & 0.264 $\pm$ 0.061 & 0.222 $\pm$ 0.028 \\
Ministral 3 14B Random & 0.356 & 0.329 $\pm$ 0.025 & 0.348 $\pm$ 0.024 & 0.338 $\pm$ 0.037 \\
Ministral 3 8B Random & 0.344 & 0.327 $\pm$ 0.043 & 0.358 $\pm$ 0.016 & 0.347 $\pm$ 0.048 \\
Ministral 3 3B kNN & 0.322 & 0.333 $\pm$ 0.022 & 0.344 $\pm$ 0.042 & 0.293 $\pm$ 0.030 \\
Gemma 4 E2B Random & 0.300 & 0.291 $\pm$ 0.021 & 0.329 $\pm$ 0.062 & 0.324 $\pm$ 0.016 \\
Gemma 4 E2B kNN+MMR & 0.289 & 0.271 $\pm$ 0.048 & 0.264 $\pm$ 0.033 & 0.216 $\pm$ 0.023 \\
Qwen 3.5 9B Random & 0.278 & 0.338 $\pm$ 0.032 & 0.362 $\pm$ 0.033 & 0.342 $\pm$ 0.029 \\
Qwen 3.5 2B Random & 0.222 & 0.242 $\pm$ 0.035 & 0.231 $\pm$ 0.039 & 0.233 $\pm$ 0.030 \\
Ministral 3 3B Random & 0.189 & 0.271 $\pm$ 0.047 & 0.296 $\pm$ 0.040 & 0.300 $\pm$ 0.014 \\
Gemma 4 E4B Random & 0.133 & 0.144 $\pm$ 0.029 & 0.147 $\pm$ 0.023 & 0.111 $\pm$ 0.014 \\
\bottomrule
\end{tabular}
\label{tab:medetec_mixed_accuracy_support_sweep}
\end{table}

\subsection{Increasing support-context size gives positive but diminishing returns}

\begin{figure}[!t]
\centering
\includegraphics[width=\linewidth]{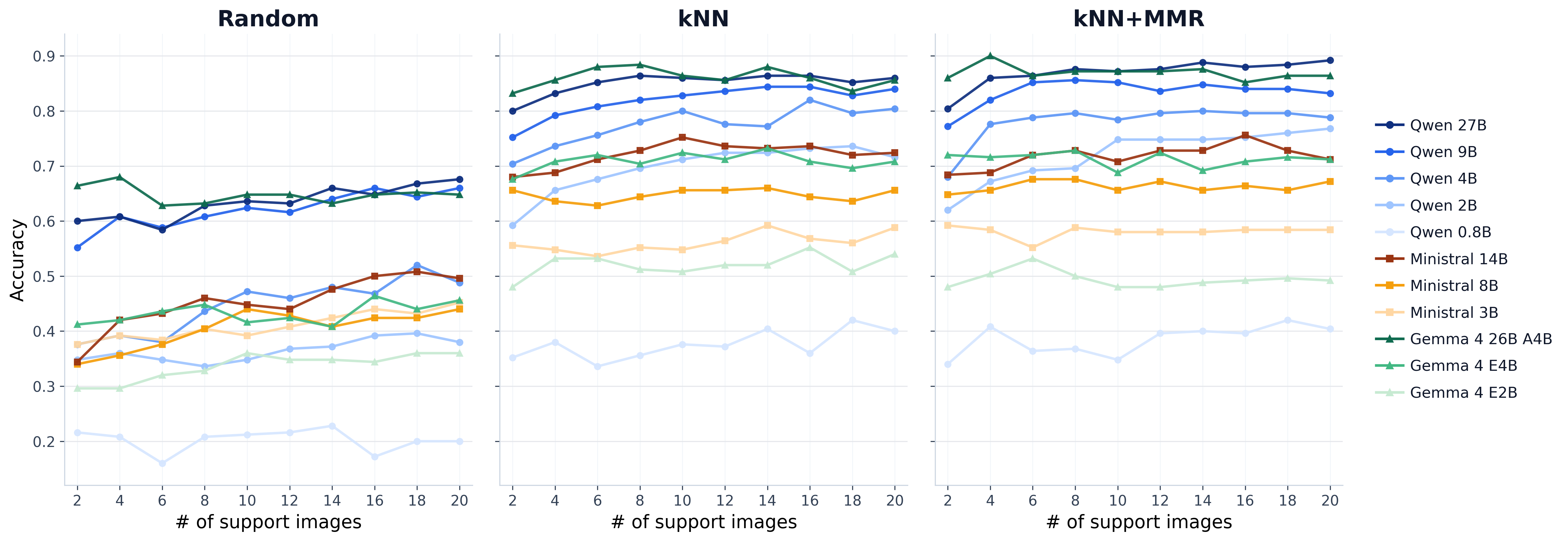}
\caption{SMLM accuracy as a function of support-context size for different support-selection strategies.}
\label{fig:support_context_accuracy_by_strategy}
\end{figure}

Figure~\ref{fig:support_context_accuracy_by_strategy} summarises the sweep over support-context sizes from 2 to 20 images. Accuracy generally increased with increasing context size, with the clearest gains in the lower part of the sweep (2-10 images). Beyond this range, performance typically flattened or fluctuated, and the largest contexts did not yield consistent additional gains: from 2 to 10 support images, mean accuracy increased by 0.012 per two-image step for $k$NN and by 0.009 per two-image step for $k$NN+MMR. From 10 to 20 support images, the corresponding gains were only 0.001 and 0.002 per step, indicating that larger contexts produced limited additional improvement.

Across the full sweep, retrieval-based context construction remained substantially stronger than random support sampling. Both $k$NN and $k$NN+MMR were consistently above the corresponding random-context results, while the difference between $k$NN and $k$NN+MMR was comparatively smaller, ranging from -0.004 to 0.020 accuracy points. The larger and mid-sized models showed clearer improvements as retrieved context was added, whereas the smallest models exhibited flatter and more variable trends.

The computational measures increased with context size. Mean response time rose with the number of support images for every model (see Table~\ref{tab:support-context-latency-by-model-support}) and mean input-token counts also increased approximately linearly with context size within each model family, from 702 to 3958 tokens for Qwen~3.5, from 843 to 4652 for Ministral~3, and from 937 to 6042 for Gemma~4 (see Table~\ref{tab:support-context-input-tokens-by-family-support}).

\begin{table}[!b]
\centering
\caption{Mean response time in seconds per request by model and number of support images. Values are averaged across the three retrieval strategies.}
\label{tab:support-context-latency-by-model-support}
\scriptsize
\setlength{\tabcolsep}{3.5pt}
\begin{tabular}{lcccccccccc}
\toprule
Model & 2 & 4 & 6 & 8 & 10 & 12 & 14 & 16 & 18 & 20 \\
\midrule
Qwen 3.5 27B & 8.36 & 12.55 & 16.49 & 21.07 & 24.68 & 28.37 & 32.27 & 36.17 & 40.38 & 44.49 \\
Qwen 3.5 9B & 3.00 & 4.94 & 6.92 & 8.68 & 10.56 & 12.23 & 13.73 & 14.10 & 15.91 & 18.17 \\
Qwen 3.5 4B & 2.06 & 3.21 & 4.33 & 5.60 & 5.53 & 5.96 & 7.21 & 8.55 & 9.57 & 10.41 \\
Qwen 3.5 2B & 0.88 & 1.43 & 1.93 & 2.51 & 2.99 & 3.44 & 3.86 & 4.30 & 4.83 & 5.28 \\
Qwen 3.5 0.8B & 0.41 & 0.65 & 0.86 & 1.14 & 1.40 & 1.64 & 1.85 & 2.11 & 2.34 & 2.49 \\
Ministral 3 14B & 4.35 & 6.78 & 9.11 & 11.71 & 14.68 & 17.57 & 20.04 & 21.97 & 24.84 & 26.95 \\
Ministral 3 8B & 2.92 & 4.62 & 6.27 & 8.16 & 9.54 & 11.11 & 12.87 & 14.57 & 16.09 & 18.40 \\
Ministral 3 3B & 1.53 & 2.46 & 3.32 & 4.24 & 5.21 & 6.13 & 7.09 & 8.00 & 9.02 & 10.13 \\
Gemma 4 26B A4B & 3.98 & 6.59 & 9.36 & 12.30 & 14.79 & 18.96 & 20.52 & 22.89 & 25.38 & 28.89 \\
Gemma 4 E4B & 2.21 & 3.61 & 5.02 & 6.33 & 7.60 & 8.93 & 10.32 & 11.73 & 13.24 & 14.67 \\
Gemma 4 E2B & 1.48 & 2.48 & 3.43 & 4.36 & 5.37 & 6.58 & 7.91 & 8.59 & 9.60 & 10.60 \\
\bottomrule
\end{tabular}
\end{table}

\begin{table}[!b]
\centering
\caption{Mean input tokens per request by model family and number of support images.}
\label{tab:support-context-input-tokens-by-family-support}
\scriptsize
\setlength{\tabcolsep}{4pt}
\begin{tabular}{lcccccccccc}
\toprule
Model family & 2 & 4 & 6 & 8 & 10 & 12 & 14 & 16 & 18 & 20 \\
\midrule
Qwen 3.5 & 702 & 1099 & 1458 & 1853 & 2226 & 2560 & 2913 & 3262 & 3626 & 3958 \\
Ministral 3 & 843 & 1305 & 1736 & 2173 & 2614 & 3020 & 3443 & 3851 & 4260 & 4652 \\
Gemma 4 & 937 & 1503 & 2070 & 2637 & 3203 & 3772 & 4339 & 4907 & 5474 & 6042 \\
\bottomrule
\end{tabular}
\end{table}

\section{Discussion}
This study demonstrates that retrieval-augmented in-context learning (ICL) improves the capabilities of small multimodal language models, making them a viable and flexible approach for wound image classification, particularly under limited labelled data. Across both tested datasets, zero-shot and random few-shot prompting were consistently weaker than query-conditioned prompting, indicating that the choice of support examples is central to effective ICL. The strongest gains came from CLIP-based $k$NN retrieval and $k$NN+MMR retrieval, especially for the larger Qwen and Gemma models, which both benefited from relevant examples and outperformed matched retrieval-only weighted $k$NN controls. This suggests that the best-performing SMLMs successfully used retrieved context within the input prompt and merged it with internal knowledge gained during pretraining. At the same time, the experiments revealed practical limits: performance depended on model scale, retrieval strategy, dataset difficulty, and support-set coverage, while increasing the number of support images beyond a moderate context size yielded diminishing accuracy gains but higher latency and longer input lengths.

Tables \ref{tab:benchmark_results} and \ref{tab:medetec_benchmark_results} demonstrate that neither retrieval strategy is optimal across model families. For the Qwen family, the $k$NN+MMR retrieval strategy yielded the best performance, whereas for the Gemma family, $k$NN was the best strategy, with no clear winner for the Ministral model family. This indicates that, for a given problem and model, different retrieval strategies should be evaluated to identify the optimal solution. Notably, larger Qwen models (9B and 27B), as well as Gemma 4 26B A4B, achieved statistically significant improvements over the baseline weighted $k$NN classifier, as indicated by McNemar tests, while having very high retention scores. On the Kaggle dataset, those models retained 93–97\% of the correct $k$NN predictions, whereas on the Medetec dataset, they retained at least 80\% of the correct $k$NN predictions. These results indicate that the models were highly capable of identifying when the $k$NN prediction is correct, without being too dependent on the retrieved context, as indicated by agreement scores (see Tables~\ref{tab:retrieval_control_knn} and~\ref{tab:medetec_retrieval_control_knn}). Particularly remarkable is the performance of Qwen 3.5 27B, which achieved a retention score of 0.952 even on the more challenging Medetec dataset. In addition to retention, the best‑performing models exhibited substantial rescue scores: approximately 0.65 on Kaggle and 0.36 on Medetec. These values suggest that the models could filter out incorrect $k$NN predictions and select the correct alternative based on their internal knowledge. The statistical significance of these improvements further supports the conclusion that the models contribute meaningful corrective behaviour rather than random variation. Another important pattern is the high agreement with at least one label present in the support set. This further indicates that the models’ corrective suggestions were strongly shaped by the contextual information provided in the support examples. While such context‑dependence may introduce a form of support‑set bias, potentially limiting performance when the support set is suboptimal, it may be advantageous when the support set is constructed through a targeted retrieval strategy rather than random sampling.

Tables~\ref{tab:mixed_accuracy_support_sweep} and~\ref{tab:medetec_mixed_accuracy_support_sweep} show that several larger retrieval-augmented SMLMs were comparable or even surpassed the classic supervised fine-tuned image classification models. This advantage was most evident with reduced support: with only 50\% or 25\% of the available support/training data, the best ICL-based SMLMs outperformed the best supervised baselines on both datasets. These results should not be interpreted as showing that ICL removes the need for labelled data: as with supervised learning, retrieval-based ICL still requires representative labelled examples, and Table~\ref{tab:subset_stats} suggests that performance decreased as embedding-space coverage declined. The practical distinction is that ICL uses these examples at inference time rather than through parameter updates. As a result, the support pool or label set can be modified without retraining a classifier, making the approach more flexible when data distributions, clinical definitions, or deployment requirements change. These results, therefore, suggest that retrieval-augmented ICL is not data-free, but it may be less data-hungry than conventional supervised training and able to remain useful with smaller labelled datasets. 

The results from the support-context sweep experiment, shown in Figure~\ref{fig:support_context_accuracy_by_strategy} and Tables~\ref{tab:support-context-latency-by-model-support} and~\ref{tab:support-context-input-tokens-by-family-support}, highlighted a trade-off between accuracy and cost when using ICL with SMLMs. Adding more support images improved performance primarily up to a moderate context size (8-10 images); beyond that, the gains were small, while input length and response time continued to increase. In practice, this favours compact retrieved contexts rather than simply larger prompts: a small set of relevant examples appears to capture most of the ICL benefit with lower latency and computational cost. The gap between retrieved and random contexts again shows that relevance matters more than context size alone: additional examples help only when they provide useful evidence for the query image.

The performance differed between the Kaggle and Medetec datasets, although the overall trends were similar. The Medetec dataset was generally harder than the Kaggle one, potentially because its support set was approximately 2.5 times smaller. However, the support-set reduction experiments show that the best SMLMs on reduced Kaggle support sets (610 and 305 images) still outperformed their counterparts on the full Medetec support set (470 images). This suggests that differences in class definitions, visual heterogeneity, or inter-class separability may also have contributed to the lower Medetec performance. 

A key limitation of these datasets for image-based classification is that some clinically distinct etiologies are grouped into a single class, reducing label specificity and increasing visual heterogeneity. For example, in the Medetec dataset, venous and arterial ulcers are merged into one category (“Venous or Arterial Ulcers”), despite having different underlying causes and characteristic appearances, which can include differences in wound shape, depth, exudate, and surrounding tissue condition \cite{medline_arterial_venous_ulcers}.  This aggregation leads to high intra-class variability, making it harder for models to learn consistent visual patterns. Conversely, other classes across both datasets may exhibit overlapping visual features despite representing different conditions, further increasing inter-class ambiguity, a known challenge in medical image classification where distinctions between categories can be subtle even for experts \cite{khanal2023investigating, wei2024deep}.  Additionally, the presence of broad or weakly defined categories (e.g., “Miscellaneous”) and potential label noise due to subjective clinical annotation can further degrade model performance and generalisation. Future work should therefore evaluate SMLMs on additional wound datasets with broader wound-type coverage, as well as on medical image-classification tasks beyond wound analysis.

During experiments, refusal rates remained low overall (0.08\% across all model outputs), despite some images containing potentially sensitive visual content. A plausible explanation is that all evaluated models were instruction-tuned: prior work shows that instruction tuning improves alignment with user intent and instruction-following behaviour in both language-only and multimodal models~\cite{chung2024scaling, liu2023visual,ouyang2022training}. In addition, all experiments were performed with the temperature parameter set to zero, i.e., a deterministic setting with reduced sampling variability relative to higher-temperature decoding. We therefore interpret the low refusal rate as consistent with stronger task compliance under instruction-tuned, deterministic inference, although isolating this effect would require a dedicated ablation. Additionally, it is important to note that in this study, we didn't use cross-validation because of the time and resource constraints. We are aware that such an approach limits generalisation, but we made sure that the evaluation sets for both datasets were randomly selected and balanced.

%\ref{tab:support-context-input-tokens-by-family-support} 

In this study, we used CLIP as the embedding model for support-set retrieval. However, many other embedding models are available, including general-purpose models such as Gemini Embeddings 2~\cite{choi2026gemini} and DINOv3~\cite{simeoni2025dinov3}, as well as models fine-tuned on medical data, such as MedCLIP~\cite{wang2022medclip} and BiomedCLIP~\cite{zhang2023biomedclip}. These alternatives may be able to provide better representations of wound images, thereby improving both context-set retrieval and ICL performance. Because our focus was on SMLMs and their interaction with ICL, evaluating the impact of different embedding models was beyond the scope of this work and is left for future research. Additionally, we used SMLMs for ICL out of the box, without fine-tuning them on the task-specific datasets. Despite this, the larger models achieved higher accuracy than the fine-tuned supervised models, suggesting that task-specific fine-tuning could further improve SMLM performance~\cite{phogat2024fine, zhou2024empirical}. An investigation of the effects of fine-tuning on general performance and ICL is planned for our future work.

The deployment of AI models in clinical environments involves a range of considerations that extend beyond algorithmic performance. In healthcare, patient data are highly sensitive, which often restricts access to comprehensive datasets and imposes stringent privacy and security requirements. Consequently, models may need to operate locally rather than through cloud‑based services, and their development must account for limited data availability, regulatory constraints, and the need for robust privacy‑preserving mechanisms. The small multimodal language models evaluated in this study demonstrate considerable potential to meet the practical constraints of on‑premises deployment in healthcare settings. The smallest variants, Qwen 3.5 0.8B, 2B, 4B, Ministral 3 3B, and Gemma 4 E2B, were the most straightforward to run locally and required modest computational resources. Mid‑sized and larger models, including Qwen 3.5 9B and 27B, Ministral 3 8B and 14B, and Gemma 4 E4B and 26B A4B, also remained feasible on consumer‑grade hardware, albeit with more careful optimisation.

%The retrieval-control comparisons point to a two-part conclusion. On this task, query-conditioned retrieval is the main driver of the improvement over zero-shot classification: even small models improve substantially once relevant support examples are added. Model size, however, determines how much additional value can be extracted from that retrieved evidence beyond a retrieval-only rule. This pattern is broadly consistent with scale-dependent differences in in-context learning discussed by Wei et al.~\cite{wei2023larger}, but our results support only the behavioral conclusion above. In particular, the large models remain strongly aligned with the retrieved neighborhood while still achieving clear gains over weighted-$k$NN, so they are not well described as pure label parroters. For the smaller models, by contrast, the data do not justify strong claims either about parroting or about ignoring the support examples. The safer interpretation is that retrieved context is helpful, but these models do not exploit it as effectively as the stronger larger models.

\section*{Acknowledgments}
The authors acknowledge the Swedish Knowledge Foundation, Jönköping University, and the industrial partners, including Mölnlycke Health Care, for financially supporting the research through the following project: ETIAI (Grant number 20230036), as part of the research and education environment SPARK at Jönköping University, Sweden.

\section*{Author contributions statement} 
G.M., O.G., J.P., and E.A. designed the study. G.M. conducted the experiments. G.M., O.G., J.P., and E.A. analysed the results. All authors contributed to the writing and reviewed the manuscript. 

\section*{Declaration of generative AI and AI-assisted technologies in the manuscript preparation process}

During the preparation of this work, the author(s) used OpenAI's GPT Image 2 in order to generate schematic wound images for Figure~\ref{fig:pipeline_schema}. These images were used only for visual illustration and were not included in the datasets, model inputs, evaluation sets, or experimental analyses. After using this tool/service, the author(s) reviewed and edited the content as needed and take(s) full responsibility for the content of the published article.

\bibliographystyle{plain}
\bibliography{references}

@article{aggarwal2021diagnostic,
  title={Diagnostic accuracy of deep learning in medical imaging: a systematic review and meta-analysis},
  author={Aggarwal, Ravi and Sounderajah, Viknesh and Martin, Guy and Ting, Daniel SW and Karthikesalingam, Alan and King, Dominic and Ashrafian, Hutan and Darzi, Ara},
  journal={NPJ digital medicine},
  volume={4},
  number={1},
  pages={65},
  year={2021},
  publisher={Nature Publishing Group UK London}
}

@article{wei2024deep,
  title={Deep learning with noisy labels in medical prediction problems: a scoping review},
  author={Wei, Yishu and Deng, Yu and Sun, Cong and Lin, Mingquan and Jiang, Hongmei and Peng, Yifan},
  journal={Journal of the American Medical Informatics Association},
  volume={31},
  number={7},
  pages={1596--1607},
  year={2024},
  publisher={Oxford University Press}
}

@inproceedings{khanal2023investigating,
  title={Investigating the impact of class-dependent label noise in medical image classification},
  author={Khanal, Bidur and Hasan, SM Kamrul and Khanal, Bishesh and Linte, Cristian A},
  booktitle={Proceedings of SPIE--the International Society for Optical Engineering},
  volume={12464},
  pages={1246437},
  year={2023}
}

@misc{medline_arterial_venous_ulcers,
  title        = {Arterial vs Venous Ulcers: What's the Difference?},
  author       = {{Medline Industries}},
  year         = {2023},
  url          = {https://www.medline.com/strategies/skin-health/distinguish-venous-ulcer-arterial-wound/},
  note         = {Accessed: 2026-06-27}
}

@article{bian2025artificial,
  title={Artificial intelligence in medical imaging: from task-specific models to large-scale foundation models},
  author={Bian, Yueyan and Li, Jin and Ye, Chuyang and Jia, Xiuqin and Yang, Qi},
  journal={Chinese Medical Journal},
  volume={138},
  number={06},
  pages={651--663},
  year={2025},
  publisher={Chinese Medical Journals Publishing House Co., Ltd. 42 Dongsi Xidajie~…}
}

@inproceedings{phogat2024fine,
  title={Fine-tuning smaller language models for question answering over financial documents},
  author={Phogat, Karmvir Singh and Puranam, Sai Akhil and Dasaratha, Sridhar and Harsha, Chetan and Ramakrishna, Shashishekar},
  booktitle={Findings of the Association for Computational Linguistics: EMNLP 2024},
  pages={10528--10548},
  year={2024}
}

@inproceedings{zhou2024empirical,
  title={An empirical study on parameter-efficient fine-tuning for multimodal large language models},
  author={Zhou, Xiongtao and He, Jie and Ke, Yuhua and Zhu, Guangyao and Guti{\'e}rrez-Basulto, V{\'\i}ctor and Pan, Jeff},
  booktitle={Findings of the Association for Computational Linguistics: ACL 2024},
  pages={10057--10084},
  year={2024}
}

@misc{choi2026gemini,
  author       = {Choi, Min and Duerig, Tom},
  title        = {Gemini Embedding 2: Our first natively multimodal embedding model},
  year         = {2026},
  month        = {Mar},
  organization = {Google},
  url          = {https://blog.google/innovation-and-ai/models-and-research/gemini-models/gemini-embedding-2/},
  note         = {The Keyword Blog. Accessed: 2026-05-12}
}

@article{simeoni2025dinov3,
  title={Dinov3},
  author={Sim{\'e}oni, Oriane and Vo, Huy V and Seitzer, Maximilian and Baldassarre, Federico and Oquab, Maxime and Jose, Cijo and Khalidov, Vasil and Szafraniec, Marc and Yi, Seungeun and Ramamonjisoa, Micha{\"e}l and others},
  journal={arXiv preprint arXiv:2508.10104},
  year={2025}
}

@article{kaplan2020scaling,
  title={Scaling laws for neural language models},
  author={Kaplan, Jared and McCandlish, Sam and Henighan, Tom and Brown, Tom B and Chess, Benjamin and Child, Rewon and Gray, Scott and Radford, Alec and Wu, Jeffrey and Amodei, Dario},
  journal={arXiv preprint arXiv:2001.08361},
  year={2020}
}

@article{hoffmann2022training,
  title={Training compute-optimal large language models},
  author={Hoffmann, Jordan and Borgeaud, Sebastian and Mensch, Arthur and Buchatskaya, Elena and Cai, Trevor and Rutherford, Eliza and Casas, DDL and Hendricks, Lisa Anne and Welbl, Johannes and Clark, Aidan and others},
  journal={arXiv preprint arXiv:2203.15556},
  volume={10},
  year={2022}
}

@article{zhong2025considerations,
  title={Considerations for patient privacy of Large Language Models in health care: scoping review},
  author={Zhong, Xiaoying and Li, Siyi and Chen, Zhao and Ge, Long and Yu, Dongdong and Wang, Shijia and You, Liangzhen and Shang, Hongcai},
  journal={Journal of Medical Internet Research},
  volume={27},
  pages={e76571},
  year={2025},
  publisher={JMIR Publications Toronto, Canada}
}

@article{barbera2025ai,
  title={AI Privacy Risks \& Mitigations--Large Language Models (LLMs)},
  author={Barbera, Isabel},
  journal={European Data Protection Board},
  pages={2025--04},
  year={2025}
}

@article{agarwal2024prompt,
  title={Prompt leakage effect and defense strategies for multi-turn llm interactions},
  author={Agarwal, Divyansh and Fabbri, Alexander R and Risher, Ben and Laban, Philippe and Joty, Shafiq and Wu, Chien-Sheng},
  journal={arXiv preprint arXiv:2404.16251},
  year={2024}
}

@article{garg2025rise,
  title={The rise of small language models in healthcare: A comprehensive survey},
  author={Garg, Muskan and Raza, Shaina and Rayana, Shebuti and Liu, Xingyi and Sohn, Sunghwan},
  journal={arXiv preprint arXiv:2504.17119},
  year={2025}
}

@article{anisuzzaman2022image,
  title={Image-based artificial intelligence in wound assessment: a systematic review},
  author={Anisuzzaman, DM and Wang, Chuanbo and Rostami, Behrouz and Gopalakrishnan, Sandeep and Niezgoda, Jeffrey and Yu, Zeyun},
  journal={Advances in Wound Care},
  volume={11},
  number={12},
  pages={687--709},
  year={2022},
  publisher={SAGE Publications Sage CA: Los Angeles, CA}
}

@article{zhang2022survey,

  title={A survey of wound image analysis using deep learning: classification, detection, and segmentation},
  author={Zhang, Ruyi and Tian, Dingcheng and Xu, Dechao and Qian, Wei and Yao, Yudong},
  journal={IEEE Access},
  volume={10},
  pages={79502--79515},
  year={2022},
  publisher={IEEE}
}

@article{oquab2023dinov2,
  title={Dinov2: Learning robust visual features without supervision},
  author={Oquab, Maxime and Darcet, Timoth{\'e}e and Moutakanni, Th{\'e}o and Vo, Huy and Szafraniec, Marc and Khalidov, Vasil and Fernandez, Pierre and Haziza, Daniel and Massa, Francisco and El-Nouby, Alaaeldin and others},
  journal={arXiv preprint arXiv:2304.07193},
  year={2023}
}

@inproceedings{eslami2023pubmedclip,
  title={Pubmedclip: How much does clip benefit visual question answering in the medical domain?},
  author={Eslami, Sedigheh and Meinel, Christoph and De Melo, Gerard},
  booktitle={Findings of the Association for Computational Linguistics: EACL 2023},
  pages={1181--1193},
  year={2023}
}

@inproceedings{wang2022medclip,
  title={Medclip: Contrastive learning from unpaired medical images and text},
  author={Wang, Zifeng and Wu, Zhenbang and Agarwal, Dinesh and Sun, Jimeng},
  booktitle={Proceedings of the 2022 Conference on Empirical Methods in Natural Language Processing},
  pages={3876--3887},
  year={2022}
}

@article{zhang2023biomedclip,
  title={Biomedclip: a multimodal biomedical foundation model pretrained from fifteen million scientific image-text pairs},
  author={Zhang, Sheng and Xu, Yanbo and Usuyama, Naoto and Xu, Hanwen and Bagga, Jaspreet and Tinn, Robert and Preston, Sam and Rao, Rajesh and Wei, Mu and Valluri, Naveen and others},
  journal={arXiv preprint arXiv:2303.00915},
  year={2023}
}

@article{choi2025evaluating,
  title={Evaluating ChatGPT-4o for ophthalmic image interpretation: from in-context learning to code-free clinical tool generation},
  author={Choi, Joon Yul and Yoo, Tae Keun},
  journal={Informatics and Health},
  volume={2},
  number={2},
  pages={158--169},
  year={2025},
  publisher={Elsevier}
}

@article{shrestha2025context,
  title={In-context learning for label-efficient cancer image classification in oncology},
  author={Shrestha, Mobina and Mandal, Bishwas and Mandal, Vishal and Shrestha, Asis and Shrestha, Amir Babu},
  journal={Informatics in Medicine Unlocked},
  pages={101683},
  year={2025},
  publisher={Elsevier}
}

@article{tabjabortesi2024machine,
  title={Machine Learning Approaches for the Image-Based Identification of Surgical Wound Infections: Systematic Review},
  author={Bortesi, Tabja and Pablo, Juan and Ranisau, Jonathan and Di, Shuang and McGillion, Michael and Rosella, Laura and Johnson, Alistair and Devereaux, P. J. and Petch, Jeremy},
  journal={Journal of Medical Internet Research},
  volume={26},
  pages={e52880},
  year={2024},
  doi={10.2196/52880}
}

@inproceedings{van2025survey,
  title={A Survey on Small Language Models},
  author={Van Nguyen, Chien and Shen, Xuan and Aponte, Ryan and Xia, Yu and Basu, Samyadeep and Hu, Zhengmian and Chen, Jian and Parmar, Mihir and Kunapuli, Sasidhar and Barrow, Joe and others},
  booktitle={Proceedings of the 15th International Conference on Recent Advances in Natural Language Processing-Natural Language Processing in the Generative AI Era},
  pages={807--821},
  year={2025}
}

@inproceedings{wang2015unified,
  title={A unified framework for automatic wound segmentation and analysis with deep convolutional neural networks},
  author={Wang, Changhan and Yan, Xinchen and Smith, Max and Kochhar, Kanika and Rubin, Marcie and Warren, Stephen M and Wrobel, James and Lee, Honglak},
  booktitle={2015 37th annual international conference of the ieee engineering in medicine and biology society (EMBC)},
  pages={2415--2418},
  year={2015},
  organization={IEEE}
}

@article{yadav2019feature,
  title={Feature extraction based machine learning for human burn diagnosis from burn images},
  author={Yadav, DP and Sharma, Ashish and Singh, Madhusudan and Goyal, Ayush},
  journal={IEEE journal of translational engineering in health and medicine},
  volume={7},
  pages={1--7},
  year={2019},
  publisher={IEEE}
}

@article{scebba2022detect,
  title={Detect-and-segment: A deep learning approach to automate wound image segmentation},
  author={Scebba, Gaetano and Zhang, Jia and Catanzaro, Sabrina and Mihai, Carina and Distler, Oliver and Berli, Martin and Karlen, Walter},
  journal={Informatics in Medicine Unlocked},
  volume={29},
  pages={100884},
  year={2022},
  publisher={Elsevier}
}

@inproceedings{abubakar2019can,
  title={Can machine learning be used to discriminate between burns and pressure ulcer?},
  author={Abubakar, Aliyu and Ugail, Hassan and Bukar, Ali Maina},
  booktitle={Proceedings of SAI intelligent systems conference},
  pages={870--880},
  year={2019},
  organization={Springer}
}

@article{goyal2020recognition,
  title={Recognition of ischaemia and infection in diabetic foot ulcers: Dataset and techniques},
  author={Goyal, Manu and Reeves, Neil D and Rajbhandari, Satyan and Ahmad, Naseer and Wang, Chuan and Yap, Moi Hoon},
  journal={Computers in biology and medicine},
  volume={117},
  pages={103616},
  year={2020},
  publisher={Elsevier}
}

@article{rostami2021multiclass,
  title={Multiclass wound image classification using an ensemble deep CNN-based classifier},
  author={Rostami, Behrouz and Anisuzzaman, DM and Wang, Chuanbo and Gopalakrishnan, Sandeep and Niezgoda, Jeffrey and Yu, Zeyun},
  journal={Computers in Biology and Medicine},
  volume={134},
  pages={104536},
  year={2021},
  publisher={Elsevier}
}

@article{aldoulah2023novel,
  title={A novel fused multi-class deep learning approach for chronic wounds classification},
  author={Aldoulah, Zaid A and Malik, Hafiz and Molyet, Richard},
  journal={Applied Sciences},
  volume={13},
  number={21},
  pages={11630},
  year={2023},
  publisher={MDPI}
}

@article{wang2020fully,
  title={Fully automatic wound segmentation with deep convolutional neural networks},
  author={Wang, Chuanbo and Anisuzzaman, DM and Williamson, Victor and Dhar, Mrinal Kanti and Rostami, Behrouz and Niezgoda, Jeffrey and Gopalakrishnan, Sandeep and Yu, Zeyun},
  journal={Scientific reports},
  volume={10},
  number={1},
  pages={21897},
  year={2020},
  publisher={Nature Publishing Group UK London}
}

@article{ramachandram2022fully,
  title={Fully automated wound tissue segmentation using deep learning on mobile devices: Cohort study},
  author={Ramachandram, Dhanesh and Ramirez-GarciaLuna, Jose Luis and Fraser, Robert DJ and Mart{\'\i}nez-Jim{\'e}nez, Mario Aurelio and Arriaga-Caballero, Jesus E and Allport, Justin},
  journal={JMIR mHealth and uHealth},
  volume={10},
  number={4},
  pages={e36977},
  year={2022},
  publisher={JMIR Publications Inc., Toronto, Canada}
}

@article{patel2024integrated,
  title={Integrated image and location analysis for wound classification: a deep learning approach},
  author={Patel, Yash and Shah, Tirth and Dhar, Mrinal Kanti and Zhang, Taiyu and Niezgoda, Jeffrey and Gopalakrishnan, Sandeep and Yu, Zeyun},
  journal={Scientific Reports},
  volume={14},
  number={1},
  pages={7043},
  year={2024},
  publisher={Nature Publishing Group UK London}
}

@article{wu2020unified,
  title={A unified framework for automatic detection of wound infection with artificial intelligence},
  author={Wu, Jin-Ming and Tsai, Chia-Jui and Ho, Te-Wei and Lai, Feipei and Tai, Hao-Chih and Lin, Ming-Tsan},
  journal={Applied Sciences},
  volume={10},
  number={15},
  pages={5353},
  year={2020},
  publisher={MDPI}
}

@article{mclean2025multimodal,
  title={Multimodal machine learning to predict surgical site infection with healthcare workload impact assessment},
  author={McLean, Kenneth A and Sgr{\`o}, Alessandro and Brown, Leo R and Buijs, Louis F and Mountain, Katie E and Shaw, Catherine A and Drake, Thomas M and Pius, Riinu and Knight, Stephen R and Fairfield, Cameron J and others},
  journal={npj Digital Medicine},
  volume={8},
  number={1},
  pages={121},
  year={2025},
  publisher={Nature Publishing Group UK London}
}

@article{zhang2024challenges,
  title={On the challenges and perspectives of foundation models for medical image analysis},
  author={Zhang, Shaoting and Metaxas, Dimitris},
  journal={Medical image analysis},
  volume={91},
  pages={102996},
  year={2024},
  publisher={Elsevier}
}

@article{lu2024small,
  title={Small language models: Survey, measurements, and insights},
  author={Lu, Zhenyan and Li, Xiang and Cai, Dongqi and Yi, Rongjie and Liu, Fangming and Zhang, Xiwen and Lane, Nicholas D and Xu, Mengwei},
  journal={arXiv preprint arXiv:2409.15790},
  year={2024}
}

@article{xu2024device,
  title={On-device language models: A comprehensive review},
  author={Xu, Jiajun and Li, Zhiyuan and Chen, Wei and Wang, Qun and Gao, Xin and Cai, Qi and Ling, Ziyuan},
  journal={arXiv preprint arXiv:2409.00088},
  year={2024}
}

@article{guan2021domain,
  title={Domain adaptation for medical image analysis: a survey},
  author={Guan, Hao and Liu, Mingxia},
  journal={IEEE Transactions on Biomedical Engineering},
  volume={69},
  number={3},
  pages={1173--1185},
  year={2021},
  publisher={IEEE}
}

@article{wang2024comprehensive,
  title={A comprehensive survey on deep active learning in medical image analysis},
  author={Wang, Haoran and Jin, Qiuye and Li, Shiman and Liu, Siyu and Wang, Manning and Song, Zhijian},
  journal={Medical Image Analysis},
  volume={95},
  pages={103201},
  year={2024},
  publisher={Elsevier}
}

@article{sun2025visual,
  title={Visual-language foundation models in medical imaging: A systematic review and meta-analysis of diagnostic and analytical applications},
  author={Sun, Yiyao and Wen, Xinran and Zhang, Yan and Jin, Lijun and Yang, Chunna and Zhang, Qianhui and Jiang, Mingchen and Xu, Zhaoyang and Guo, Wei and Su, Juan and others},
  journal={Computer Methods and Programs in Biomedicine},
  volume={268},
  pages={108870},
  year={2025},
  publisher={Elsevier}
}

@inproceedings{radford2021clip,
  title={Learning transferable visual models from natural language supervision},
  author={Radford, Alec and Kim, Jong Wook and Hallacy, Chris and Ramesh, Aditya and Goh, Gabriel and Agarwal, Sandhini and Sastry, Girish and Askell, Amanda and Mishkin, Pamela and Clark, Jack and others},
  booktitle={International conference on machine learning},
  pages={8748--8763},
  year={2021},
  organization={PmLR}
}

@article{ferber2024context,
  title={In-context learning enables multimodal large language models to classify cancer pathology images},
  author={Ferber, Dyke and W{\"o}lflein, Georg and Wiest, Isabella C and Ligero, Marta and Sainath, Srividhya and Ghaffari Laleh, Narmin and El Nahhas, Omar SM and M{\"u}ller-Franzes, Gustav and J{\"a}ger, Dirk and Truhn, Daniel and others},
  journal={Nature Communications},
  volume={15},
  number={1},
  pages={10104},
  year={2024},
  publisher={Nature Publishing Group UK London}
}

@article{brown2020language,
  title={Language models are few-shot learners},
  author={Brown, Tom and Mann, Benjamin and Ryder, Nick and Subbiah, Melanie and Kaplan, Jared D and Dhariwal, Prafulla and Neelakantan, Arvind and Shyam, Pranav and Sastry, Girish and Askell, Amanda and others},
  journal={Advances in neural information processing systems},
  volume={33},
  pages={1877--1901},
  year={2020}
}

@article{hu2022lora,
  title={Lora: Low-rank adaptation of large language models.},
  author={Hu, Edward J and Shen, Yelong and Wallis, Phillip and Allen-Zhu, Zeyuan and Li, Yuanzhi and Wang, Shean and Wang, Liang and Chen, Weizhu and others},
  journal={Iclr},
  volume={1},
  number={2},
  pages={3},
  year={2022}
}

@article{dettmers2023qlora,
  title={Qlora: Efficient finetuning of quantized llms},
  author={Dettmers, Tim and Pagnoni, Artidoro and Holtzman, Ari and Zettlemoyer, Luke},
  journal={Advances in neural information processing systems},
  volume={36},
  pages={10088--10115},
  year={2023}
}

@article{dosovitskiy2021image,
  title={An Image is Worth 16x16 Words: Transformers for Image Recognition at Scale},
  author={Dosovitskiy, Alexey and Beyer, Lucas and Kolesnikov, Alexander and Weissenborn, Dirk and Zhai, Xiaohua and Unterthiner, Thomas and Dehghani, Mostafa and Minderer, Matthias and Heigold, Georg and Gelly, Sylvain and Uszkoreit, Jakob and Houlsby, Neil},
  journal={arXiv preprint arXiv:2010.11929},
  year={2021}
}

@inproceedings{liu2022convnet,
  title={A ConvNet for the 2020s},
  author={Liu, Zhuang and Mao, Hanzi and Wu, Chao-Yuan and Feichtenhofer, Christoph and Darrell, Trevor and Xie, Saining},
  booktitle={Proceedings of the IEEE/CVF Conference on Computer Vision and Pattern Recognition},
  pages={11976--11986},
  year={2022}
}

@article{tan2021efficientnetv2,
  title={EfficientNetV2: Smaller Models and Faster Training},
  author={Tan, Mingxing and Le, Quoc V},
  journal={arXiv preprint arXiv:2104.00298},
  year={2021}
}

@inproceedings{he2016deep,
  title={Deep Residual Learning for Image Recognition},
  author={He, Kaiming and Zhang, Xiangyu and Ren, Shaoqing and Sun, Jian},
  booktitle={Proceedings of the IEEE Conference on Computer Vision and Pattern Recognition},
  pages={770--778},
  year={2016}
}

@article{nori2023can,
  title={Can generalist foundation models outcompete special-purpose tuning? case study in medicine},
  author={Nori, Harsha and Lee, Yin Tat and Zhang, Sheng and Carignan, Dean and Edgar, Richard and Fusi, Nicolo and King, Nicholas and Larson, Jonathan and Li, Yuanzhi and Liu, Weishung and others},
  journal={arXiv preprint arXiv:2311.16452},
  year={2023}
}

@article{zhang2023makes,
  title={What makes good examples for visual in-context learning?},
  author={Zhang, Yuanhan and Zhou, Kaiyang and Liu, Ziwei},
  journal={Advances in Neural Information Processing Systems},
  volume={36},
  pages={17773--17794},
  year={2023}
}

@inproceedings{li2024configure,
  title={How to configure good in-context sequence for visual question answering},
  author={Li, Li and Peng, Jiawei and Chen, Huiyi and Gao, Chongyang and Yang, Xu},
  booktitle={Proceedings of the IEEE/CVF Conference on Computer Vision and Pattern Recognition},
  pages={26710--26720},
  year={2024}
}

@inproceedings{chen2025can,
  title={Can Multimodal Large Language Models Truly Perform Multimodal In-Context Learning?},
  author={Chen, Shuo and Han, Zhen and He, Bailan and Liu, Jianzhe and Buckley, Mark and Qin, Yao and Torr, Philip and Tresp, Volker and Gu, Jindong},
  booktitle={2025 IEEE/CVF Winter Conference on Applications of Computer Vision (WACV)},
  pages={6000--6010},
  year={2025},
  organization={IEEE}
}

@article{suo2024visual,
  title={Visual prompt selection for in-context learning segmentation},
  author={Suo, Wei and Lai, Lanqing and Sun, Mengyang and Zhang, Hanwang and Wang, Peng and Zhang, Yanning},
  journal={arXiv preprint arXiv:2407.10233},
  year={2024}
}

@article{lee2026learning,
  title={Learning to Select Visual In-Context Demonstrations},
  author={Lee, Eugene and Lin, Yu-Chi and Diao, Jiajie},
  journal={arXiv preprint arXiv:2603.26775},
  year={2026}
}

@inproceedings{luo2024does,
  title={How does the textual information affect the retrieval of multimodal in-context learning?},
  author={Luo, Yang and Zheng, Zangwei and Zhu, Zirui and You, Yang},
  booktitle={Proceedings of the 2024 Conference on Empirical Methods in Natural Language Processing},
  pages={5321--5335},
  year={2024}
}

@inproceedings{dong2024survey,
  title={A survey on in-context learning},
  author={Dong, Qingxiu and Li, Lei and Dai, Damai and Zheng, Ce and Ma, Jingyuan and Li, Rui and Xia, Heming and Xu, Jingjing and Wu, Zhiyong and Chang, Baobao and others},
  booktitle={Proceedings of the 2024 conference on empirical methods in natural language processing},
  pages={1107--1128},
  year={2024}
}

@inproceedings{moor2023med,
  title={Med-flamingo: a multimodal medical few-shot learner},
  author={Moor, Michael and Huang, Qian and Wu, Shirley and Yasunaga, Michihiro and Dalmia, Yash and Leskovec, Jure and Zakka, Cyril and Reis, Eduardo Pontes and Rajpurkar, Pranav},
  booktitle={Machine learning for health (ML4H)},
  pages={353--367},
  year={2023},
  organization={PMLR}
}

@article{nam2025multimodal,
  title={Multimodal large language models in medical imaging: current state and future directions},
  author={Nam, Yoojin and Kim, Dong Yeong and Kyung, Sunggu and Seo, Jinyoung and Song, Jeong Min and Kwon, Jimin and Kim, Jihyun and Jo, Wooyoung and Park, Hyungbin and Sung, Jimin and others},
  journal={Korean Journal of Radiology},
  volume={26},
  number={10},
  pages={900},
  year={2025}
}

@article{bradshaw2025large,
  title={Large language models and large multimodal models in medical imaging: a primer for physicians},
  author={Bradshaw, Tyler J and Tie, Xin and Warner, Joshua and Hu, Junjie and Li, Quanzheng and Li, Xiang},
  journal={Journal of nuclear medicine},
  volume={66},
  number={2},
  pages={173--182},
  year={2025},
  publisher={Society of Nuclear Medicine}
}

@article{tsimpoukelli2021multimodal,
  title={Multimodal few-shot learning with frozen language models},
  author={Tsimpoukelli, Maria and Menick, Jacob L and Cabi, Serkan and Eslami, SM and Vinyals, Oriol and Hill, Felix},
  journal={Advances in Neural Information Processing Systems},
  volume={34},
  pages={200--212},
  year={2021}
}

@article{jiang2024many,
  title={Many-shot in-context learning in multimodal foundation models},
  author={Jiang, Yixing and Irvin, Jeremy and Wang, Ji Hun and Chaudhry, Muhammad Ahmed and Chen, Jonathan H and Ng, Andrew Y},
  journal={arXiv preprint arXiv:2405.09798},
  year={2024}
}

@inproceedings{li2023blip,
  title={Blip-2: Bootstrapping language-image pre-training with frozen image encoders and large language models},
  author={Li, Junnan and Li, Dongxu and Savarese, Silvio and Hoi, Steven},
  booktitle={International conference on machine learning},
  pages={19730--19742},
  year={2023},
  organization={PMLR}
}

@article{sun2025medical,
  title={Medical multimodal foundation models in clinical diagnosis and treatment: Applications, challenges, and future directions},
  author={Sun, Kai and Xue, Siyan and Sun, Fuchun and Sun, Haoran and Luo, Yu and Wang, Ling and Wang, Siyuan and Guo, Na and Liu, Lei and Zhao, Tian and others},
  journal={Artificial Intelligence in Medicine},
  pages={103265},
  year={2025},
  publisher={Elsevier}
}

@article{ayhan2025context,
  title={In-context learning for data-efficient diabetic retinopathy detection via multimodal foundation models},
  author={Ayhan, Murat S and Ong, Ariel Y and Ruffell, Eden and Wagner, Siegfried K and Merle, David A and Keane, Pearse A},
  journal={Ophthalmology Science},
  pages={100934},
  year={2025},
  publisher={Elsevier}
}

@incollection{fisher1970statistical,
  title={Statistical methods for research workers},
  author={Fisher, Ronald Aylmer},
  booktitle={Breakthroughs in statistics: Methodology and distribution},
  pages={66--70},
  year={1970},
  publisher={Springer}
}

@article{azad2023foundational,
  title={Foundational models in medical imaging: A comprehensive survey and future vision},
  author={Azad, Bobby and Azad, Reza and Eskandari, Sania and Bozorgpour, Afshin and Kazerouni, Amirhossein and Rekik, Islem and Merhof, Dorit},
  journal={arXiv preprint arXiv:2310.18689},
  year={2023}
}

@article{baharoon2023evaluating,
  title={Evaluating general purpose vision foundation models for medical image analysis: An experimental study of dinov2 on radiology benchmarks},
  author={Baharoon, Mohammed and Qureshi, Waseem and Ouyang, Jiahong and Xu, Yanwu and Aljouie, Abdulrhman and Peng, Wei},
  journal={arXiv preprint arXiv:2312.02366},
  year={2023}
}

@article{anisuzzaman2022multimodal,
  title={Multi-modal wound classification using wound image and location by deep neural network},
  author={Anisuzzaman, D M and Rostami, Behnoosh and Niezgoda, Joseph and Gopalakrishna, Ashwin and Yu, Zhihong and Liang, Zhiqiang},
  journal={Scientific Reports},
  volume={12},
  number={1},
  pages={17298},
  year={2022},
  doi={10.1038/s41598-022-21813-0}
}

@article{alayrac2022flamingo,
  title={Flamingo: a visual language model for few-shot learning},
  author={Alayrac, Jean-Baptiste and Donahue, Jeff and Luc, Pauline and Miech, Antoine and Barr, Iain and Hasson, Yana and Lenc, Karel and Mensch, Arthur and Millican, Katherine and Reynolds, Malcolm and others},
  journal={Advances in neural information processing systems},
  volume={35},
  pages={23716--23736},
  year={2022}
}

@inproceedings{carbonell1998use,
  title={The use of MMR, diversity-based reranking for reordering documents and producing summaries},
  author={Carbonell, Jaime and Goldstein, Jade},
  booktitle={Proceedings of the 21st annual international ACM SIGIR conference on Research and development in information retrieval},
  pages={335--336},
  year={1998}
}

@misc{qwen,
    title  = {{Qwen3.5}: Towards Native Multimodal Agents},
    author = {{Qwen Team}},
    month  = {February},
    year   = {2026},
    url    = {https://qwen.ai/blog?id=qwen3.5}
}

@article{mistral2026ministral3,
  title   = {Ministral 3},
  author  = {{Mistral AI}},
  journal = {arXiv preprint arXiv:2601.08584},
  year    = {2026},
  url     = {https://arxiv.org/abs/2601.08584},
  doi     = {10.48550/arXiv.2601.08584}
}

@misc{gemma_2026_g4,
  title        = {Gemma 4: Our most capable open models to date},
  author       = {{Gemma Team, Google DeepMind}},
  howpublished = {Google Blog},
  year         = {2026},
  url          = {https://blog.google/innovation-and-ai/technology/developers-tools/gemma-4/},
  note         = {Accessed: 2026-04-15}
}

@article{ouyang2022training,
  title={Training language models to follow instructions with human feedback},
  author={Ouyang, Long and Wu, Jeffrey and Jiang, Xu and Almeida, Diogo and Wainwright, Carroll and Mishkin, Pamela and Zhang, Chong and Agarwal, Sandhini and Slama, Katarina and Ray, Alex and others},
  journal={Advances in neural information processing systems},
  volume={35},
  pages={27730--27744},
  year={2022}
}

@article{chung2024scaling,
  title={Scaling instruction-finetuned language models},
  author={Chung, Hyung Won and Hou, Le and Longpre, Shayne and Zoph, Barret and Tay, Yi and Fedus, William and Li, Yunxuan and Wang, Xuezhi and Dehghani, Mostafa and Brahma, Siddhartha and others},
  journal={Journal of Machine Learning Research},
  volume={25},
  number={70},
  pages={1--53},
  year={2024}
}

@article{liu2023visual,
  title={Visual instruction tuning},
  author={Liu, Haotian and Li, Chunyuan and Wu, Qingyang and Lee, Yong Jae},
  journal={Advances in neural information processing systems},
  volume={36},
  pages={34892--34916},
  year={2023}
}

@article{mcnemar1947note,
  title={Note on the sampling error of the difference between correlated proportions or percentages},
  author={McNemar, Quinn},
  journal={Psychometrika},
  volume={12},
  number={2},
  pages={153--157},
  year={1947},
  publisher={Springer-Verlag}
}

@article{cohen1960coefficient,
  title={A coefficient of agreement for nominal scales},
  author={Cohen, Jacob},
  journal={Educational and psychological measurement},
  volume={20},
  number={1},
  pages={37--46},
  year={1960},
  publisher={Sage Publications Sage CA: Thousand Oaks, CA}
}

@inproceedings{zhao2021calibrate,
  title={Calibrate before use: Improving few-shot performance of language models},
  author={Zhao, Zihao and Wallace, Eric and Feng, Shi and Klein, Dan and Singh, Sameer},
  booktitle={International conference on machine learning},
  pages={12697--12706},
  year={2021},
  organization={Pmlr}
}

@inproceedings{min2022rethinking,
  title={Rethinking the role of demonstrations: What makes in-context learning work?},
  author={Min, Sewon and Lyu, Xinxi and Holtzman, Ari and Artetxe, Mikel and Lewis, Mike and Hajishirzi, Hannaneh and Zettlemoyer, Luke},
  booktitle={Proceedings of the 2022 conference on empirical methods in natural language processing},
  pages={11048--11064},
  year={2022}
}

@article{xiong2026retrieving,
  title={Retrieving Counterfactuals Improves Visual In-Context Learning},
  author={Xiong, Guangzhi and Sinha, Sanchit and He, Zhenghao and Zhang, Aidong},
  journal={arXiv preprint arXiv:2603.16737},
  year={2026}
}

@inproceedings{kapuriya2025exploring,
  title={Exploring the role of diversity in example selection for in-context learning},
  author={Kapuriya, Janak and Kaushik, Manit and Ganguly, Debasis and Bhatia, Sumit},
  booktitle={Proceedings of the 48th International ACM SIGIR Conference on Research and Development in Information Retrieval},
  pages={2962--2966},
  year={2025}
}

\newpage

\appendix

\section{Retrieval-control association statistics}
\label{sec:supp_retrieval_control_association}

To complement the paired retrieval-control comparisons, we report conditional correctness and association statistics for each model--control comparison for both Kaggle and Medetec datasets. For each paired $2\times2$ correctness table, we computed the retention rate, $P(\mathrm{SMLM~correct}\mid\mathrm{control~correct})$, the rescue rate, $P(\mathrm{SMLM~correct}\mid\mathrm{control~wrong})$, Fisher's exact test $p$-value, and the corresponding odds ratio (OR).

\begin{table*}[!ht]
\centering
\caption{Conditional correctness and association statistics for $k$NN few-shot prompting relative to the global weighted-$k$NN control in the Kaggle dataset. Ret.\ denotes $P(\mathrm{SMLM\ correct}\mid\mathrm{control\ correct})$, Resc.\ denotes $P(\mathrm{SMLM\ correct}\mid\mathrm{control\ wrong})$, Fisher $p$ is the exact Fisher test $p$-value, and OR denotes the corresponding odds ratio.}
\label{tab:supp_conditional_knn}
\small
\setlength{\tabcolsep}{5pt}
\begin{tabular}{lcccc}
\toprule
\textbf{Model} & \textbf{Ret.} & \textbf{Resc.} & \textbf{Fisher $p$} & \textbf{OR} \\
\midrule
Gemma~4 26B A4B & 0.966 & 0.630 & $1.81 \times 10^{-11}$ & 16.73 \\
Qwen~3.5 27B    & 0.960 & 0.616 & $1.86 \times 10^{-11}$ & 15.11 \\
Qwen~3.5 9B     & 0.938 & 0.589 & $1.56 \times 10^{-10}$ & 10.53 \\
Qwen~3.5 4B     & 0.921 & 0.438 & $1.11 \times 10^{-15}$ & 14.92 \\
Ministral~3 14B & 0.881 & 0.438 & $1.36 \times 10^{-12}$ & 9.52 \\
Gemma~4 E4B     & 0.876 & 0.397 & $5.10 \times 10^{-14}$ & 10.69 \\
Qwen~3.5 2B     & 0.853 & 0.342 & $5.18 \times 10^{-15}$ & 11.15 \\
Ministral~3 8B  & 0.785 & 0.342 & $6.66 \times 10^{-11}$ & 7.02 \\
Ministral~3 3B  & 0.633 & 0.342 & $4.26 \times 10^{-5}$  & 3.31 \\
Gemma~4 E2B     & 0.520 & 0.301 & 0.002 & 2.51 \\
Qwen~3.5 0.8B   & 0.362 & 0.247 & 0.103 & 1.73 \\
\bottomrule
\end{tabular}
\end{table*}

\begin{table*}[!ht]
\centering
\caption{Conditional correctness and association statistics for $k$NN+MMR few-shot prompting relative to the matched weighted-$k$NN control recomputed on the final reranked support set in the Kaggle dataset. Ret.\ denotes $P(\mathrm{SMLM\ correct}\mid\mathrm{control\ correct})$, Resc.\ denotes $P(\mathrm{SMLM\ correct}\mid\mathrm{control\ wrong})$, Fisher $p$ is the exact Fisher test $p$-value, and OR denotes the corresponding odds ratio.}
\label{tab:supp_conditional_mmr}
\small
\setlength{\tabcolsep}{5pt}
\begin{tabular}{lcccc}
\toprule
\textbf{Model} & \textbf{Ret.} & \textbf{Resc.} & \textbf{Fisher $p$} & \textbf{OR} \\
\midrule
Gemma~4 26B A4B & 0.941 & 0.629 & $1.17 \times 10^{-8}$  & 9.49 \\
Qwen~3.5 27B    & 0.936 & 0.645 & $1.01 \times 10^{-7}$  & 8.07 \\
Qwen~3.5 9B     & 0.926 & 0.694 & $1.45 \times 10^{-5}$  & 5.49 \\
Qwen~3.5 4B     & 0.904 & 0.435 & $2.33 \times 10^{-13}$ & 12.24 \\
Qwen~3.5 2B     & 0.856 & 0.419 & $6.80 \times 10^{-11}$ & 8.26 \\
Ministral~3 14B & 0.814 & 0.387 & $1.11 \times 10^{-9}$  & 6.92 \\
Ministral~3 8B  & 0.761 & 0.339 & $3.76 \times 10^{-9}$  & 6.20 \\
Gemma~4 E4B     & 0.851 & 0.306 & $2.55 \times 10^{-15}$ & 12.93 \\
Ministral~3 3B  & 0.654 & 0.355 & $5.12 \times 10^{-5}$  & 3.44 \\
Gemma~4 E2B     & 0.484 & 0.323 & 0.028 & 1.97 \\
Qwen~3.5 0.8B   & 0.340 & 0.177 & 0.016 & 2.39 \\
\bottomrule
\end{tabular}
\end{table*}

\begin{table*}[!t]
\centering
\caption{Conditional correctness and association statistics for $k$NN few-shot prompting relative to the global weighted-$k$NN control in the Medetec dataset. Ret.\ denotes $P(\mathrm{SMLM\ correct}\mid\mathrm{control\ correct})$, Resc.\ denotes $P(\mathrm{SMLM\ correct}\mid\mathrm{control\ wrong})$, Fisher $p$ is the exact Fisher test $p$-value, and OR denotes the corresponding odds ratio.}
\label{tab:medetec_supp_conditional_knn}
\small
\setlength{\tabcolsep}{5pt}
\begin{tabular}{lcccc}
\toprule
\textbf{Model} & \textbf{Ret.} & \textbf{Resc.} & \textbf{Fisher $p$} & \textbf{OR} \\
\midrule
Qwen~3.5 27B        & 0.952 & 0.354 & $9.35 \times 10^{-10}$ & 36.47 \\
Qwen~3.5 9B         & 0.857 & 0.375 & $3.71 \times 10^{-6}$  & 10.00 \\
Gemma~4 26B A4B     & 0.810 & 0.354 & $1.61 \times 10^{-5}$  & 7.75 \\
Qwen~3.5 4B         & 0.810 & 0.375 & $3.93 \times 10^{-5}$  & 7.08 \\
Ministral~3 14B     & 0.714 & 0.312 & $2.79 \times 10^{-4}$  & 5.50 \\
Gemma~4 E4B         & 0.905 & 0.208 & $9.60 \times 10^{-12}$ & 36.10 \\
Qwen~3.5 2B         & 0.714 & 0.375 & 0.002 & 4.17 \\
Ministral~3 8B      & 0.690 & 0.229 & $1.60 \times 10^{-5}$  & 7.50 \\
Gemma~4 E2B         & 0.524 & 0.229 & 0.005 & 3.70 \\
Ministral~3 3B      & 0.476 & 0.188 & 0.006 & 3.94 \\
Qwen~3.5 0.8B       & 0.214 & 0.271 & 0.626 & 0.73 \\
\bottomrule
\end{tabular}
\end{table*}

\begin{table*}[!ht]
\centering
\caption{Conditional correctness and association statistics for $k$NN+MMR few-shot prompting relative to the matched weighted-$k$NN control recomputed on the final reranked support set in the Medetec dataset. Ret.\ denotes $P(\mathrm{SMLM\ correct}\mid\mathrm{control\ correct})$, Resc.\ denotes $P(\mathrm{SMLM\ correct}\mid\mathrm{control\ wrong})$, Fisher $p$ is the exact Fisher test $p$-value, and OR denotes the corresponding odds ratio.}
\label{tab:medetec_supp_conditional_mmr}
\small
\setlength{\tabcolsep}{5pt}
\begin{tabular}{lcccc}
\toprule
\textbf{Model} & \textbf{Ret.} & \textbf{Resc.} & \textbf{Fisher $p$} & \textbf{OR} \\
\midrule
Qwen~3.5 27B        & 0.976 & 0.429 & $8.58 \times 10^{-9}$  & 53.33 \\
Qwen~3.5 9B         & 0.951 & 0.327 & $2.70 \times 10^{-10}$ & 40.22 \\
Gemma~4 26B A4B     & 0.878 & 0.388 & $1.50 \times 10^{-6}$  & 11.37 \\
Qwen~3.5 4B         & 0.829 & 0.449 & $2.20 \times 10^{-4}$  & 5.96 \\
Ministral~3 14B     & 0.805 & 0.245 & $1.83 \times 10^{-7}$  & 12.72 \\
Gemma~4 E4B         & 0.854 & 0.224 & $1.45 \times 10^{-9}$  & 20.15 \\
Qwen~3.5 2B         & 0.780 & 0.388 & $2.49 \times 10^{-4}$  & 5.61 \\
Ministral~3 8B      & 0.756 & 0.204 & $1.81 \times 10^{-7}$  & 12.09 \\
Ministral~3 3B      & 0.610 & 0.204 & $1.00 \times 10^{-4}$  & 6.09 \\
Gemma~4 E2B         & 0.463 & 0.143 & 0.001 & 5.18 \\
Qwen~3.5 0.8B       & 0.220 & 0.184 & 0.793 & 1.25 \\
\bottomrule
\end{tabular}
\end{table*}

\clearpage
\newpage

\section{Detailed kNN+MMR retrieval-control results}
\label{sec:supp_mmr_retrieval_control}

This appendix reports the full paired retrieval-control results for the $k$NN+MMR prompting condition. These results are discussed only briefly in Section~\ref{sec:results_retrieval_control} because the diagnostic logic is identical to the primary $k$NN analysis in the main text.

\begin{table*}[!ht]
\centering
\caption{Paired retrieval-control comparisons for $k$NN+MMR few-shot prompting against the matched weighted-$k$NN control recomputed on the final reranked support set (accuracy = 0.752) on the Kaggle dataset. Top-1, Maj., and Any denote match rates to the top retrieved support label, the support-majority label, and any label present in the retrieved support set, respectively. Ret.\ denotes $P(\mathrm{SMLM\ correct}\mid\mathrm{control\ correct})$, and Resc.\ denotes $P(\mathrm{SMLM\ correct}\mid\mathrm{control\ wrong})$. Agreement denotes the raw prediction agreement rate with the control, and $\kappa$ denotes Cohen's kappa.}
\label{tab:retrieval_control_mmr}
\scriptsize
\setlength{\tabcolsep}{3pt}
\begin{tabular}{lcccccccccc}
\toprule
\textbf{Model} & \textbf{Acc.} & \textbf{$\Delta$} & \textbf{McNemar $p$} & \textbf{Top-1} & \textbf{Maj.} & \textbf{Any} & \textbf{Ret.} & \textbf{Resc.} & \textbf{Agreement} & \textbf{$\kappa$} \\
\midrule
Qwen~3.5 9B      & 0.868 & +0.116 & $<0.001$ & 0.704 & 0.756 & 0.972 & 0.926 & 0.694 & 0.744 & 0.715 \\
Qwen~3.5 27B     & 0.864 & +0.112 & $<0.001$ & 0.700 & 0.768 & 0.984 & 0.936 & 0.645 & 0.760 & 0.733 \\
Gemma~4 26B A4B  & 0.864 & +0.112 & $<0.001$ & 0.692 & 0.764 & 0.984 & 0.941 & 0.629 & 0.752 & 0.724 \\
Qwen~3.5 4B      & 0.788 & +0.036 & 0.233    & 0.680 & 0.768 & 0.984 & 0.904 & 0.435 & 0.768 & 0.741 \\
Qwen~3.5 2B      & 0.748 & -0.004 & 1.000    & 0.708 & 0.740 & 0.948 & 0.856 & 0.419 & 0.736 & 0.705 \\
Ministral~3 14B  & 0.708 & -0.044 & 0.193    & 0.640 & 0.688 & 0.920 & 0.814 & 0.387 & 0.672 & 0.634 \\
Ministral~3 8B   & 0.656 & -0.096 & 0.004    & 0.632 & 0.648 & 0.916 & 0.761 & 0.339 & 0.644 & 0.602 \\
Gemma~4 E4B      & 0.716 & -0.036 & 0.243    & 0.684 & 0.736 & 0.972 & 0.851 & 0.306 & 0.724 & 0.691 \\
Ministral~3 3B   & 0.580 & -0.172 & $<0.001$ & 0.540 & 0.548 & 0.820 & 0.654 & 0.355 & 0.536 & 0.484 \\
Gemma~4 E2B      & 0.444 & -0.308 & $<0.001$ & 0.396 & 0.408 & 0.668 & 0.484 & 0.323 & 0.400 & 0.332 \\
Qwen~3.5 0.8B    & 0.300 & -0.452 & $<0.001$ & 0.284 & 0.292 & 0.636 & 0.340 & 0.177 & 0.292 & 0.205 \\
\bottomrule
\end{tabular}
\end{table*}

%Under $k$NN+MMR few-shot prompting on the Medetec dataset, the matched weighted-$k$NN control reached 0.456 accuracy. In this setting, more models significantly exceeded the retrieval-only control then under $k$NN retrieval. Qwen~3.5 27B achieved the strongest result, reaching 0.678 accuracy and improving over the matched control by +0.222. Qwen~3.5 9B, Gemma~4 26B A4B, and Qwen~3.5 4B also significantly exceeded the control, with gains of +0.156, +0.156, and +0.167, respectively. These models again showed high retention of control-correct cases, ranging from 0.829 to 0.976, and somewhat substantial rescue of control errors, ranging from 0.327 to 0.449. Fisher exact tests indicated strong positive association with the matched control for all four models (odds ratios $=5.96$--$53.33$; see Appendix~\ref{sec:supp_retrieval_control_association}). However, their agreement and $\kappa$ values were lower than those observed for the strongest models on the Kaggle dataset, indicating that the Medetec dataset was more difficult retrieval-conditioned decision setting.

%The weaker models again failed to outperform the matched retrieval-only control. Ministral~3 8B was exactly at parity with the control, while Qwen~3.5 0.8B and Gemma~4 E2B fell significantly below it. Although several of these smaller models still improved over their zero-shot and random few-shot baselines, their paired comparisons show that those gains were largely explained by the retrieved examples themselves rather than by reliable multimodal reasoning beyond retrieval-only classification.

\begin{table*}[!ht]
\centering
\caption{Paired retrieval-control comparisons for $k$NN+MMR few-shot prompting against the matched weighted-$k$NN control recomputed on the final reranked support set (control accuracy = 0.456) on the Medetec dataset. Top-1, Maj., and Any denote match rates to the top retrieved support label, the support-majority label, and any label present in the retrieved support set, respectively. Ret.\ denotes $P(\mathrm{SMLM\ correct}\mid\mathrm{control\ correct})$, and Resc.\ denotes $P(\mathrm{SMLM\ correct}\mid\mathrm{control\ wrong})$. Agreement denotes the raw prediction agreement rate with the control, and $\kappa$ denotes Cohen's kappa.}
\label{tab:medetec_retrieval_control_mmr}
\scriptsize
\setlength{\tabcolsep}{3pt}
\begin{tabular}{lcccccccccc}
\toprule
\textbf{Model} & \textbf{Acc.} & \textbf{$\Delta$} & \textbf{McNemar $p$} & \textbf{Top-1} & \textbf{Maj.} & \textbf{Any} & \textbf{Ret.} & \textbf{Resc.} & \textbf{Agreement} & \textbf{$\kappa$} \\
\midrule
Qwen~3.5 27B        & 0.678 & +0.222 & $<0.001$ & 0.600 & 0.667 & 0.956 & 0.976 & 0.429 & 0.667 & 0.589 \\
Qwen~3.5 9B         & 0.611 & +0.156 & 0.001    & 0.522 & 0.589 & 0.978 & 0.951 & 0.327 & 0.589 & 0.522 \\
Gemma~4 26B A4B     & 0.611 & +0.156 & 0.007    & 0.522 & 0.567 & 0.933 & 0.878 & 0.388 & 0.567 & 0.488 \\
Qwen~3.5 4B         & 0.622 & +0.167 & 0.008    & 0.411 & 0.500 & 0.844 & 0.829 & 0.449 & 0.500 & 0.419 \\
Ministral~3 14B     & 0.500 & +0.044 & 0.503    & 0.533 & 0.578 & 0.967 & 0.805 & 0.245 & 0.578 & 0.492 \\
Gemma~4 E4B         & 0.511 & +0.056 & 0.332    & 0.511 & 0.600 & 0.933 & 0.854 & 0.224 & 0.600 & 0.505 \\
Qwen~3.5 2B         & 0.567 & +0.111 & 0.087    & 0.456 & 0.433 & 0.911 & 0.780 & 0.388 & 0.433 & 0.379 \\
Ministral~3 8B      & 0.456 & +0.000 & 1.000    & 0.522 & 0.511 & 0.856 & 0.756 & 0.204 & 0.511 & 0.411 \\
Ministral~3 3B      & 0.389 & -0.067 & 0.327    & 0.300 & 0.356 & 0.767 & 0.610 & 0.204 & 0.356 & 0.290 \\
Gemma~4 E2B         & 0.289 & -0.167 & 0.008    & 0.333 & 0.278 & 0.844 & 0.463 & 0.143 & 0.267 & 0.178 \\
Qwen~3.5 0.8B       & 0.200 & -0.256 & $<0.001$ & 0.167 & 0.111 & 0.433 & 0.220 & 0.184 & 0.111 & 0.060 \\
\bottomrule
\end{tabular}
\end{table*}

\newpage

\end{document}